\documentclass{article} 

\usepackage{amsmath,amsfonts,bm}

\def\eqref#1{equation~\ref{#1}}

\def\1{\bm{1}}

\DeclareMathAlphabet{\mathsfit}{\encodingdefault}{\sfdefault}{m}{sl}
\SetMathAlphabet{\mathsfit}{bold}{\encodingdefault}{\sfdefault}{bx}{n}

\usepackage{iclr2026_conference,times}

\title{Visual CoT Makes VLMs Smarter but More Fragile}

\author{Chunxue Xu$^{\dagger}$,\quad
Yiwei Wang$^{\mathsection}$, \quad
Yujun Cai$^{\diamond}$, \quad
Bryan Hooi$^{\ddagger}$, \quad
Songze Li$^{\dagger}$\thanks{Corresponding Author.} 
\\
$^{\dagger}$ Southeast University, China 
\quad $^{\mathsection}$ University of California, Merced, USA \\
 $^{\diamond}$ The University of Queensland, Australia 
\quad $^{\ddagger}$ National University of Singapore, Singapore
}

\iclrfinalcopy 
\begin{document}

\maketitle

\begin{abstract}

Chain-of-Thought (CoT) techniques have significantly enhanced reasoning in Vision-Language Models (VLMs). Extending this paradigm, Visual CoT integrates explicit visual edits, such as cropping or annotating regions of interest, into the reasoning process, achieving superior multimodal performance. However, the robustness of Visual CoT-based VLMs against image-level noise
remains unexplored. In this paper, we present the first systematic evaluation of Visual CoT robustness under visual perturbations. Our benchmark spans 12 image corruption types across 4 Visual Question Answering (VQA) datasets, enabling a comprehensive comparison between VLMs that use Visual CoT, and VLMs that do not. The results reveal that integrating Visual CoT consistently improves absolute accuracy regardless of whether the input images are clean or corrupted by noise; however, it also increases sensitivity to input perturbations, resulting in sharper performance degradation compared to standard VLMs.
Through extensive analysis, we identify the intermediate reasoning components of Visual CoT, i.e., the edited image patches
, as the primary source of fragility. Building on this analysis, we propose a plug-and-play robustness enhancement method that integrates Grounding DINO model into the Visual CoT pipeline, providing high-confidence local visual cues to stabilize reasoning.
Our work reveals clear fragility patterns in Visual CoT and offers an effective, architecture-agnostic solution for enhancing visual robustness.
\end{abstract}

\section{Introduction}
With the introduction of Chain-of-Thought (CoT) techniques, Large Language Models (LLMs) have achieved remarkable progress in reasoning capabilities. Recent studies have extended CoT to Vision-Language Models (VLMs), evolving from CoT pipelines that rely solely on textual reasoning to Visual Chain-of-Thought (Visual CoT) approaches that incorporate visual information into the reasoning process \citep{shao2024visual, jiang2025mme, wang2025visuothink, Chen_Zhou_Shen_2024, Fu2025ReFocusVE}, thereby significantly enhancing multimodal reasoning performance.
For Visual CoT methods, a common practice is to perform visual editing on the input images, such as cropping, annotating, or modifying regions of interest. The edited and original images are jointly fed into the model, which is guided to perform step-by-step reasoning based on both visual inputs, thus supporting finer-grained multimodal understanding.

\begin{figure}
    \centering
    \includegraphics[width=1\linewidth]{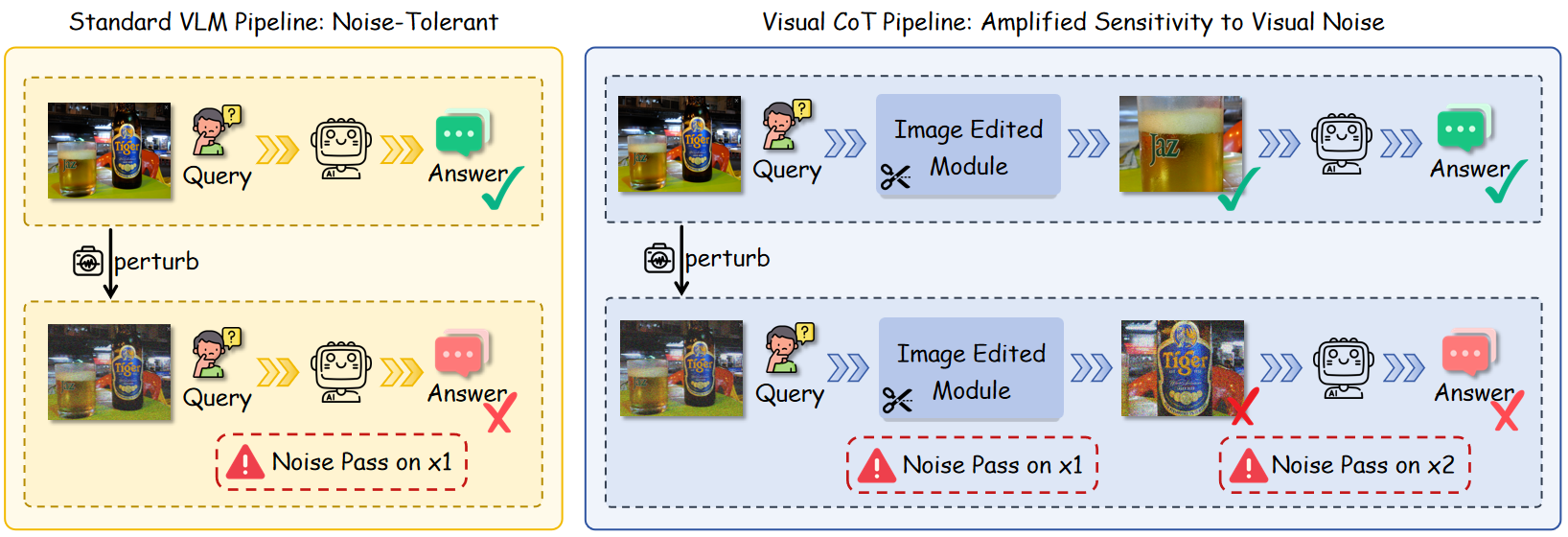}
    \caption{Visual CoT pipeline amplifies the effects of input noise due to intermediate visual editing steps, where noise influences both global and local components, in contrast to Standard VLMs where noise only affects a single input stage.}

    \label{fig:perform_2_par}
\end{figure}

However, despite several studies exploring the robustness of CoT-based VLMs under purely textual reasoning scenarios \citep{jiang2025mme, zhou2024can, jiang2025misaligning, wang2024stop}, there has been no systematic investigation into the robustness of Visual CoT-based VLMs. Unlike standard textual CoT, Visual CoT introduces explicit visual manipulations, which inherently interact with the reasoning process. Under noisy conditions, these added components may amplify the effects of input perturbations, making the system more susceptible to errors and raising new challenges for multimodal reasoning (as shown in \Cref{fig:perform_2_par}).

To address this gap, we propose a robustness evaluation framework for Visual CoT-based VLMs, aiming to systematically assess how Visual CoT affects model robustness under visual perturbations. Specifically, we employ 12 distinct visual perturbation techniques, systematically applied to the input images. 
To quantify the robustness impact introduced by Visual CoT, we compute the performance degradation separately under two paradigms, Visual CoT-enhanced VLMs and standard VLMs without CoT, by comparing outputs before and after perturbation. This setup enables a direct and systematic analysis of how incorporating Visual CoT changes robustness in multimodal reasoning tasks.


Our experimental results show that while Visual CoT VLMs consistently achieve higher absolute accuracy across all perturbation conditions, their performance degrades more sharply under noisy inputs compared to standard VLMs. Further experiments reveal that noise compromises the reliability of intermediate reasoning components, which subsequently propagates to the final stage and leads to more severe accuracy degradation.
Moreover, attention analysis demonstrates that Visual CoT VLMs exhibit a more concentrated attention distribution compared to standard VLMs. 
These attention characteristics explain why Visual CoT VLMs achieve higher accuracy under whether clean or perturbed inputs.

Building on this analysis, we propose a lightweight, plug-and-play robustness enhancement strategy by integrating Grounding DINO model \citep{liu2023grounding} into the Visual CoT pipeline. Grounding DINO automatically identifies high-confidence image regions relevant to the question and injects these localized cues into the reasoning chain. 
This provides the model with richer and redundant visual information, effectively mitigating the adverse effects of perturbation without requiring additional fine-tuning or architecture modification.

This paper makes several contributions to the literature:
\circled{1}We present the first comprehensive evaluation of Visual CoT-based VLMs under visual perturbations, offering new insights into the trade-off between their accuracy and robustness.
\circled{2}We introduce a lightweight strategy that integrates Grounding DINO into the Visual CoT pipeline, 
substantially improving reasoning stability under noisy conditions.
\circled{3}We establish a robustness evaluation benchmark that spans 12 visual perturbation types across multiple datasets on VQA task, providing a reproducible and extensible foundation for future research on Visual CoT robustness.

\section{Related Work}
\subsection{Chain-of-Thought in Vision-Language Models}
CoT prompting has significantly advanced the reasoning capabilities of LLMs by decomposing complex problems into intermediate steps. Building on this success, recent research has extended CoT to VLMs, enabling multimodal reasoning by integrating visual evidence into the reasoning chain. \cite{zhang2024cocot} first formally introduce the concept of Multimodal-CoT (MCoT) and extend it into a rationalizing-answering stages paradigm. 
\cite{yang2023mm} introduce MM-REACT, which combines LLMs with vision experts through prompt-based coordination, enabling zero-shot multimodal reasoning across diverse visual tasks.

Later, the development of CoT gradually evolved from purely textual prompting to frameworks that integrate explicit visual editing into the reasoning process. \cite{shao2024visual} introduce a large-scale visual CoT dataset with bounding boxes and reasoning steps, enabling \vlm to identify key regions and enhance multimodal reasoning.
\cite{hu2024visual} propose Sketchpad, enabling \vlm to sketch visual artifacts during reasoning and improving performance on complex visual and mathematical tasks.
\cite{Fu2025ReFocusVE} introduce ReFous, which equips \vlm with visual editing capabilities to generate “visual thoughts”, achieving substantial gains in structured image understanding.
These studies highlight the progression of CoT in VLMs, from textual prompting to Visual CoT pipelines incorporating editing and sketching, broadening the scope and effectiveness of multimodal reasoning.

\subsection{Robustness of Chain-of-Thought}
While CoT has achieved remarkable performance gains, its robustness under input perturbations remains an open challenge.
\cite{zhou_can_2024} address the challenge of noisy rationales in CoT prompting by introducing the NoRa dataset and proposing CD-CoT, a contrastive denoising method that significantly improves reasoning robustness under irrelevant or inaccurate intermediate thoughts.
\cite{jin2024saber} investigate the security vulnerabilities of CoT-based models in code generation and propose SABER, a model-agnostic backdoor attack leveraging self-attention, demonstrating that CoT models remain highly susceptible to stealthy data poisoning.
\cite{wang2024stop} reveal that CoT-based MLLMs exhibit only limited resistance to adversarial attacks despite their multi-step reasoning process.

Collectively, these studies reveal that despite the strong reasoning capabilities of CoT-based systems, they remain vulnerable to various forms of security threats.
However, prior studies have predominantly investigated robustness in purely textual CoT prompting frameworks, the robustness of Visual CoT approaches remains largely underexplored.
In this work, we aim to bridge this gap by conducting a comprehensive study on the robustness of Visual CoT reasoning under diverse noisy and adversarial conditions.

\section{Methodology}
To systematically examine the impact of incorporating Visual CoT on model robustness, we compare two paradigms: (1) Standard VLMs, which generate answers directly from image–question pairs, and (2) Visual CoT VLMs, which explicitly integrate intermediate visual reasoning steps. Our evaluation focuses on how their performance degrades when subjected to identical perturbations applied to the input images.

\subsection{Two VLM Paradigms}
In this study, we compare two modeling paradigms:

\textbf{1) Standard VLMs\ }
The Standard VLMs refers to the conventional multimodal question answering framework, where the model receives an image–question pair and directly generates the final answer. The reasoning process is entirely implicit within the model’s internal representations, without interpretable intermediate steps. 

\textbf{2) Visual CoT VLMs\ }  
In contrast, Visual CoT VLMs introduce intermediate reasoning stages into the multimodal pipeline. We adopt VisCoT \citep{shao2024visual} as a representative implementation. Given an image–question pair, VisCoT first predicts a bounding box for the most relevant region, and then crops the corresponding patch. This patch and the original image are jointly encoded, and their visual features are fused with the textual input to produce the final answer.


\label{sec:perturb-method}
\subsection{Perturbation Design}
To evaluate the robustness of Standard and Visual CoT VLMs under noisy conditions, we design controlled perturbation experiments by applying both natural corruptions and adversarial attacks to the visual inputs and measuring model performance degradation across different perturbation types.

For natural perturbations, we adopt image-level corruption strategies from the ImageNet-C benchmark \citep{HendrycksD19}, which cover a broad range of common distortions. We select 8 types grouped into four categories:
(1) Noise: Gaussian Noise, Shot Noise, Impulse Noise;
(2) Blur: Defocus Blur, Zoom Blur;
(3) Digital: Pixelate, Elastic Transformations, Contrast Adjustments.
Each corruption is applied at 5 severity levels, enabling fine-grained analysis of performance degradation under increasing noise intensity. The detailed description about severity setting is shown in \Cref{severity_natural}.

In addition, we incorporate widely used white-box adversarial attacks, including FGSM \citep{Goodfellow2014ExplainingAH}, BIM \citep{kurakin2017adversarial} , PGD \citep{DBLP:conf/iclr/MakelovLN25} and C\&W \citep{Carlini2016TowardsET}. 
In our white-box attack setup, the generation of adversarial examples for Visual CoT VLMs follows the same strategy as for Standard VLMs. Specifically, we only apply adversarial perturbations to the initial input image, without modifying intermediate localized patches. This ensures a fair comparison between the two paradigms by keeping the attack surface and perturbation budget consistent.
We also manually define 5 severity levels for each attack based on parameters (e.g., iteration count, step size) that determine the strength and impact of the perturbation.
The detailed implementation procedures of these algorithms, along with the generated adversarial examples, are provided in the \Cref{severity_White_box}.

\subsection{Robustness Evaluation Metrics}
We adopt Visual Question Answering (VQA) as the primary evaluation task, as it is a representative and widely used benchmark for assessing multimodal reasoning capabilities.  
Formally, the VQA task is defined as follows:  

\noindent
Given a natural language question $q$ and an associated image $i$ as the visual context, the model is required to generate an answer.  
Each question is paired with a ground truth answer $gt$, which serves as the reference for evaluation.  
Our evaluation dataset $\mathcal{D}_{\text{eval}}$ consists of triplets $(q, i, gt)$.  
For a given Vision-Language Model $f$ that takes $(q, i)$ as input and outputs an answer $f(q, i)$, we define the \textbf{Answer Accuracy} over $\mathcal{D}_{\text{eval}}$ as:  
\begin{equation*}
\text{Acc}(f, \mathcal{D}_{\text{eval}}) \overset{def}{=} \frac{1}{\left | \mathcal{D}_{\text{eval}}\right |}\sum_{(q,i,gt)\in \mathcal{D}_{\text{eval}}}\mathbbm{1}(f(q,i), gt),
\label{eq:acc}
\end{equation*}
\noindent
where $\mathbbm{1}(\cdot)$ is the indicator function, returning 1 if $f(q, i)$ exactly matches $gt$, and 0 otherwise. 
In our evaluation, the indicator function is implemented by \textsc{GPT-4o} acting as an automatic evaluator, which compares the predicted and ground truth answers while accounting for minor paraphrasing or synonym variations. 

To evaluate model robustness, we apply input-level perturbations during inference by directly corrupting the image. Formally, given a perturbation operator $\delta(\cdot)$, the perturbed evaluation set is defined as
$\delta(\mathcal{D}_{\text{eval}}) = \left\{ (\delta(q_k), i_k, gt_k)\right\}_{k=1}^{N}$,
where $\delta$ is applied to the image in each sample of the original dataset $\mathcal{D}_{\text{eval}}$.
Following      \citet{zhu2024promptrobustevaluatingrobustnesslarge}, we quantify the relative performance degradation caused by perturbations using the \textbf{Performance Drop Rate (PDR)}:  
\begin{equation*}
\text{PDR}(f) \overset{def}{=} 
\frac{
\text{Acc}(f, \mathcal{D}_{\text{eval}})
- 
\text{Acc}(f, \delta(\mathcal{D}_{\text{eval}}))
}
{
\text{Acc}(f, \mathcal{D}_{\text{eval}})
},
\end{equation*}
\noindent
where $\delta(\mathcal{D}_{\text{eval}})$ denotes the evaluation set in which perturbations are applied to the input image before being processed by the model.
A higher PDR value indicates greater performance degradation under noise, while a lower value suggests stronger robustness.

\section{Experiments}
\subsection{Evaluation Models and Evaluation Datasets}
We conduct a comprehensive robustness evaluation of the two paradigms, Standard VLM and Visual CoT VLM, under input perturbations. We use two representative \vlm, \llavamodel \citep{liu2023llava} and 
\viscotmodel \citep{shao2024visual}, each evaluated under both paradigms by toggling the use of Visual CoT reasoning.

Our evaluation spans four widely adopted datasets across both natural and document-based VQA tasks: \cub \citep{Wah2011TheCB}, \sroie \citep{DBLP:conf/icdar/HuangCHBKLJ19}, \docvqa \citep{mathew2021docvqa}, and \textcaps \citep{sidorov2020textcaps}. 
For each dataset, we construct corresponding perturbed variants by applying the image perturbation techniques described in Section~\ref{sec:perturb-method}, resulting in 48 image perturbation evaluation splits (e.g., CUB-Gaussian\_Noise, CUB-Shot\_Noise, SROIE-Gaussian\_Noise, SROIE-Shot\_Noise, etc.).


\subsection{Evaluation Results}
As summarized in \Cref{tab:perturbimg_pdr} and \Cref{tab:perturbimg_acc}, our evaluation reveals distinct robustness characteristics between Standard VLMs and their Visual CoT-enhanced counterparts when subjected to image corruptions: 

(1) \viscot exhibit a higher PDR than \pvqa in 70 out of 96 evaluated settings. Specifically, the average PDR of \viscot reaches 26.3\%, while that of \pvqa is only 18.6\%. This indicates that \viscot are generally more vulnerable to perturbations compared to \pvqa. The trend is particularly pronounced on the \textcaps dataset, where \viscot exhibit higher performance degradation than the \pvqa in 100\% of perturbation cases.

(2) Although \viscot exhibit lower robustness than \pvqa in terms of PDR, their accuracy under perturbations remains higher in 79 out of 96 cases. Notably, on the \cub and \textcaps datasets, \pvqa occasionally show an apparent accuracy increase under perturbations; however, even in these cases, the resulting accuracy still falls below that of the perturbed \viscot.

\begin{table*}[t]
\centering
\renewcommand{\arraystretch}{1.2}
\caption{PDR (\%) of two VLM paradigms (\textbf{Standard VLMs} vs. \textbf{Visual CoT VLMs}) under 12 image perturbations across four datasets and two base models (\llavamodel, \viscotmodel).}
\label{tab:robustness}
\resizebox{\textwidth}{!}{
\begin{tabular}{c|c|c|cccccccccccc}
\toprule
\textbf{Dataset} & \textbf{Model} & \textbf{Paradigm} & \textbf{Gaussian} & \textbf{Shot} & \textbf{Impulse} & \textbf{Defocus} & \textbf{Zoom} & \textbf{Pixelate} & \textbf{Elastic} & \textbf{Contrast} &
\textbf{BIM} & 
\textbf{FGSM} & 
\textbf{PGD} & 
\textbf{C\&W} \\ 
\midrule

\multirow{4}{*}{\textbf{\cub}}
 & \multirow{2}{*}{\llavamodel} 
   & Standard     & -10.3 	&-3.4 	&-3.4 	&-10.3 	&-13.8 	&\textbf{3.4} 	&-10.3 	&-10.3 &10.3 &10.3 &\textbf{6.9} &\textbf{13.8} 
 \\
 &         & VisCoT & \textbf{13.2} 	&\textbf{5.3} 	&\textbf{10.5} 	&\textbf{7.9} 	&\textbf{5.3} 	&-2.6 	&\textbf{13.2} 	&\textbf{23.7} &\textbf{10.5} &\textbf{10.5} &2.6 &13.1
 \\
 \cmidrule(lr){2-15}
 & \multirow{2}{*}{\viscotmodel} 
   & Standard     & 0.0 	&-6.9 	&-10.3 	&-10.3 	&-13.8 	&\textbf{3.4} 	&-3.4 	&-10.3 &3.4 & 1.0 &\textbf{6.8} &3.4
    \\
 &         & Visual CoT & \textbf{2.6} 	&\textbf{2.6} 	&\textbf{10.5} 	&\textbf{7.9} 	&\textbf{5.3} 	&-2.6 	&\textbf{10.5} 	&\textbf{23.7} 
&\textbf{13.1} &\textbf{15.7} &2.6 &\textbf{15.8}
    \\ \midrule

\multirow{4}{*}{\textbf{\sroie}}
 & \multirow{2}{*}{\llavamodel} 
   & Standard     & 11.1 	&\textbf{22.2} 	&-11.1 	&0.0 	&77.8 	&\textbf{44.4} 	&-88.9 	&7.3 &\textbf{77.8} &\textbf{77.8} &66.7 &\textbf{77.8}
    \\
 &         & Visual CoT & \textbf{27.3} 	&18.2 	&\textbf{12.1} 	&\textbf{9.4} 	&\textbf{90.9} 	&9.1 	&\textbf{9.1} 	&\textbf{34.8} &68.5 &75.8 &\textbf{72.7} &73.9 
    \\ \cmidrule(lr){2-15}
 & \multirow{2}{*}{\viscotmodel} 
   & Standard     & \textbf{30.0} 	&\textbf{40.0} 	&\textbf{0.0} 	&10.0 	&\textbf{50.0} 	&\textbf{40.0} 	&-20.0 	&60.0 &\textbf{90.0} & \textbf{80.0} &30.0 &\textbf{96.0}
    \\
 &         & Visual CoT & 25.2 	&16.5 	&-0.7 	&\textbf{19.4} 	&28.1 	&10.8 	&\textbf{13.7} 	&\textbf{65.5} &82.7 &65.4 &\textbf{30.9} &94.2
    \\ \midrule

\multirow{4}{*}{\textbf{\docvqa}}
 & \multirow{2}{*}{\llavamodel} 
   & Standard     & \textbf{35.3} 	&\textbf{11.8} 	&-11.8 	&23.5 	&\textbf{58.8} 	&-5.9 	&29.4 	&\textbf{29.4} &41.2 &\textbf{35.2} &\textbf{47.0} &17.6
    \\
 &         & Visual CoT & -25.7 	&4.8 	&\textbf{-7.6} 	&\textbf{23.8} 	&38.1 	&\textbf{9.5} 	&\textbf{50.0} 	&4.8 &\textbf{41.9} &23.8 &33.3 &\textbf{40.9}
    \\ \cmidrule(lr){2-15}
 & \multirow{2}{*}{\viscotmodel} 
   & Standard     & \textbf{39.1} 	&\textbf{13.0} 	&-4.3 	&\textbf{47.8} 	&56.5 	&\textbf{4.3} 	&13.0 	&\textbf{47.8} &30.4 &7.8 &\textbf{56.5} &25.2
    \\
 &         & Visual CoT & -10.2 	&9.1 	&\textbf{1.9} 	&42.1 	&\textbf{56.6} 	&-9.4 	&\textbf{29.4} 	&26.0 &\textbf{54.3} &\textbf{52.8} &47.1 &\textbf{73.2}
    \\ \midrule

\multirow{4}{*}{\textbf{\textcaps}}
 & \multirow{2}{*}{\llavamodel} 
   & Standard     & 3.8 	&7.7 	&13.5 	&30.8 	&\textbf{71.9} 	&7.7 	&17.3 	&\textbf{30.8} &19.2 &25.0 &63.4 &15.4
    \\
 &         & Visual CoT & \textbf{18.0} 	&\textbf{15.6} 	&\textbf{22.7} 	&\textbf{49.0} 	&64.3 	&\textbf{14.8} 	&\textbf{32.6} 	&25.3 &\textbf{32.6} &\textbf{39.9} &\textbf{69.1} &\textbf{28.5}
    \\ \cmidrule(lr){2-15}
 & \multirow{2}{*}{\viscotmodel} 
   & Standard     & 1.8 	&2.5 	&-1.8 	&-12.5 	&-16.1 	&4.3 	&-3.6 	&8.9 &33.9 &33.9 &69.6 &35.7
    \\
 &         & Visual CoT & \textbf{25.0} 	&\textbf{13.2} 	&\textbf{20.6} 	&\textbf{10.3} 	&\textbf{14.7} 	&\textbf{7.4} 	&\textbf{17.6} 	&\textbf{20.6}  &\textbf{39.7} &\textbf{44.8} &\textbf{70.5} &\textbf{44.1}
   \\ 
 \bottomrule
\end{tabular}}
\label{tab:perturbimg_pdr}
\end{table*}

\begin{table*}[t]
\centering
\renewcommand{\arraystretch}{1.2}
\caption{Answer Accuracy (\%) of two VLM paradigms (\textbf{Standard VLMs} vs. \textbf{Visual CoT VLMs}) under 12 image perturbations across four datasets and two base models (\llavamodel, \viscotmodel).}
\resizebox{\textwidth}{!}{
\begin{tabular}{c|c|c|ccccccccccccc}
\toprule
\textbf{Dataset} & \textbf{Model} & \textbf{Paradigm} & \textbf{\textit{Clean}}&
\textbf{Gaussian} & \textbf{Shot} & \textbf{Impulse} & \textbf{Defocus} & \textbf{Zoom} & \textbf{Pixelate} & \textbf{Elastic} & \textbf{Contrast} & 
\textbf{BIM} & 
\textbf{FGSM} & 
\textbf{PGD} & 
\textbf{C\&W} \\ 
\midrule

\multirow{4}{*}{\textbf{\cub}}
 & \multirow{2}{*}{\llavamodel} 
   & Standard     & 58.0 & 64.0 & 60.0 & 60.0 & 64.0 & 66.0 & 56.0 & 64.0 & \textbf{64.0} & 64.0 & 64.0 & 54.0 & 66.0 \\
 &         & VisCoT & \textbf{76.0} & \textbf{66.0} & \textbf{72.0} & \textbf{68.0} & \textbf{70.0} & \textbf{72.0} & \textbf{78.0} & \textbf{66.0} & 58.0 & \textbf{68.0} & \textbf{68.0} & \textbf{74.0} & \textbf{66.0} \\
 \cmidrule(lr){2-16}
 & \multirow{2}{*}{\viscotmodel} 
   & Standard     & 58.0	& 58.0	& 62.0	& 64.0	& 64.0	& 66.0	& 56.0	& 60.0	 & \textbf{64.0} & 56.0 &	58.0&	54.0&	56.0 \\
 &         & Visual CoT & \textbf{76.0}	& \textbf{74.0}	& \textbf{74.0}	& \textbf{68.0}	& \textbf{70.0}	& \textbf{72.0}	& \textbf{78.0}	& \textbf{68.0}	& 58.0 & \textbf{66.0}	&\textbf{64.0}	&\textbf{74.0}	&\textbf{64.0}
    \\ \midrule

\multirow{4}{*}{\textbf{\sroie}}
 & \multirow{2}{*}{\llavamodel} 
   & Standard     & 9.0	&8.0	&7.0	&10.0	&9.0	&2.0	&5.0	&17.0	&8.3 & 2.0	&2.0	&3.0	&2.0
 \\
 &         & Visual CoT & \textbf{33.0}	&\textbf{24.0}	&\textbf{27.0}	&\textbf{29.0}	&\textbf{29.9}	&\textbf{30.0}	&\textbf{30.0}	&\textbf{30.0}	&\textbf{21.5} & \textbf{10.4}	&\textbf{8.0}	&\textbf{9.0}	&\textbf{8.6}
 \\
 \cmidrule(lr){2-16}
 & \multirow{2}{*}{\viscotmodel} 
   & Standard     & 10.0	&7.0	&6.0	&10.0	&9.0	&5.0	&6.0	&12.0	&4.0 & 1.0	&2.0	&7.0	&0.4
 \\
 &         & Visual CoT & \textbf{34.8}	&\textbf{26.0}	&\textbf{29.0}	&\textbf{35.0}	&\textbf{28.0}	&\textbf{25.0}	&\textbf{31.0}	&\textbf{30.0}	&\textbf{12.0} & \textbf{6.0}	&\textbf{12.0}	&\textbf{24.0}	&\textbf{2.0}
    \\ \midrule

\multirow{4}{*}{\textbf{\docvqa}}
 & \multirow{2}{*}{\llavamodel} 
   & Standard     & 17.0	&11.0	&15.0	&19.0	&13.0	&7.0	&18.0	&\textbf{12.0} & 12.0 & 10.0	&11.0&	9.0	&\textbf{14.0}
 \\
 &         & Visual CoT & \textbf{21.0}	&\textbf{26.4}	&\textbf{20.0}	&\textbf{22.6}	&\textbf{16.0}	&\textbf{13.0}	&\textbf{19.0}	&10.5 &\textbf{20.0} & \textbf{12.2}	&\textbf{16.0}	&\textbf{14.0}	&12.4 \\
 \cmidrule(lr){2-16}
 & \multirow{2}{*}{\viscotmodel} 
   & Standard     & 11.5	&7.0	&10.0	&12.0	&6.0	&5.0	&11.0	&10.0	&6.0 & 8.0	&10.6&	5.0	&\textbf{8.6}
 \\
 &         & Visual CoT & \textbf{26.5}	&\textbf{29.2}	&\textbf{24.1}	&\textbf{26.0}	&\textbf{15.3}	&\textbf{11.5}	&\textbf{29.0}	&\textbf{18.7}	&\textbf{19.6} & \textbf{12.1}	&\textbf{12.5}	&\textbf{14.0}	&7.1
    \\ \midrule

\multirow{4}{*}{\textbf{\textcaps}}
 & \multirow{2}{*}{\llavamodel} 
   & Standard     & 52.0	&50.0	&48.0	&45.0	&\textbf{36.0}	&14.6	&48.0	&\textbf{43.0}	&36.0 & \textbf{42.0}	&\textbf{39.0}	&19.0	&44.0 \\
 &         & Visual CoT & \textbf{61.6}	&\textbf{50.5}	&\textbf{52.0}	&\textbf{47.6}	&31.4	&\textbf{22.0}	&\textbf{52.5}	&41.5	&\textbf{46.0} & 41.5	&37.0	&19.0	&44.0 \\
 \cmidrule(lr){2-16}
 & \multirow{2}{*}{\viscotmodel} 
   & Standard     & 56.0	&\textbf{55.0}	&54.6	&\textbf{57.0}	&\textbf{63.0}	&\textbf{65.0}	&53.6	&\textbf{58.0}	&51.0 & 37.0	&37.0	&17.0	&36.0 \\
 &         & Visual CoT & \textbf{68.0}	&51.0	&\textbf{59.0}	&54.0	&61.0	&58.0	&\textbf{63.0}	&56.0	&\textbf{54.0} & \textbf{41.0}	&\textbf{37.5}	&\textbf{20.0}	&\textbf{38.0}
   \\ 
 \bottomrule
\end{tabular}}
\label{tab:perturbimg_acc}
\end{table*}

\subsection{Performance Trends Across Perturbation Severity Levels}
\begin{figure}[t]
    \centering
    \includegraphics[width=0.6\linewidth]{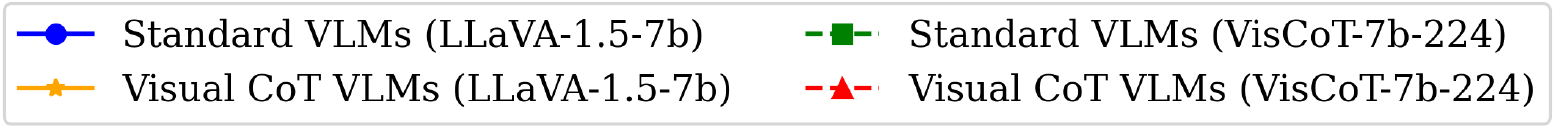}

    \vspace{0.1cm} 

    \includegraphics[width=1\linewidth]{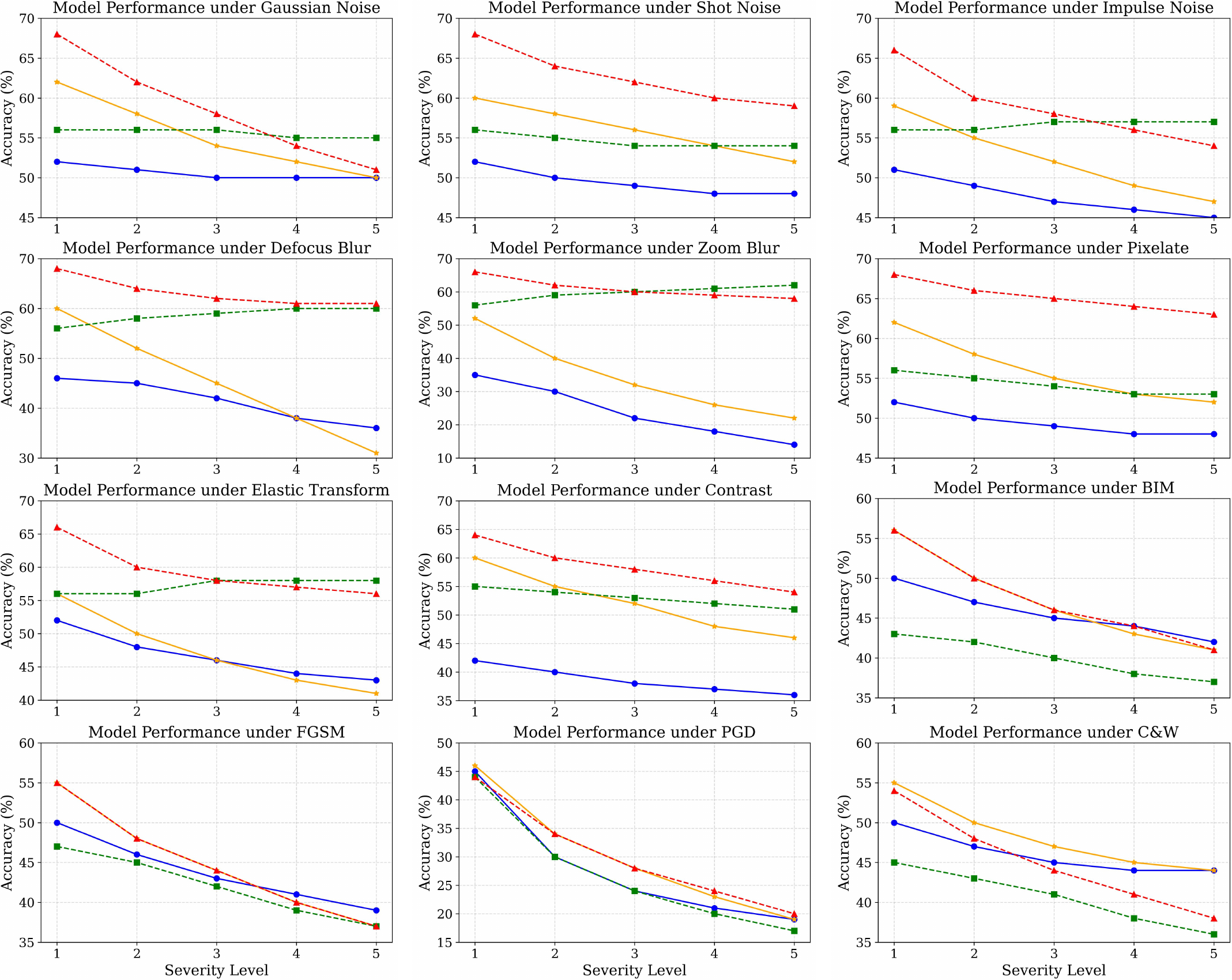}

    \caption{Accuracy trends of \viscot and \pvqa under varying image perturbation severity levels.}
    \label{fig:severity_down}
\end{figure}

To explore how model accuracy evolves under increasing levels of image corruption, we conduct a severity-aware evaluation across 5 levels of perturbation intensity. \Cref{fig:severity_down} presents representative accuracy degradation curves across all perturbation types on the \cub dataset using the \llavamodel model.

As illustrated, the \viscot' performance curve exhibits a steeper decline compared to \pvqa' as noise severity increases. 
This indicates that \viscot are more sensitive to perturbation, with performance dropping more rapidly under increasingly severe corruption. However, despite this higher degradation rate, \viscot typically maintain a  higher absolute accuracy across all severity levels. This suggests that while they are less robust, their overall capacity for accurate reasoning remains stronger than that of the \pvqa.
This pattern aligns with our previous quantitative findings on PDR and perturbed accuracy.

\section{Analysis}
\subsection{Why Are \viscot More Fragile?}

The experimental results reveal a consistent pattern: although \viscot achieve higher absolute accuracy than \pvqa baseline, they suffer a more severe accuracy drop under the same perturbation conditions. We attribute this to fundamental differences in their reasoning paradigms.

\pvqa adopt a single-step inference paradigm, directly mapping the input image and question to an answer, which reduces noise propagation and makes them less sensitive to perturbations. In contrast, \viscot employ a multi-stage chain-of-thought framework that integrates global and localized visual information. This enhances reasoning precision but also increases complexity, causing input perturbations to cascade through multiple steps and lead to sharper accuracy degradation.

To investigate, we further analyze the relationship between the accuracy of reasoning steps and the final model performance. In particular, we measure the quality of intermediate bounding box predictions using Intersection over Union (IoU, as described in \Cref{iou}), and examine how variations in IoU relate to the PDR of the overall system. The results (as shown in \Cref{{fig:iou_drop}}) reveal a clear positive correlation: when intermediate localization accuracy decreases, the final prediction accuracy also drops more severely. This indicates that errors accumulated in the intermediate reasoning stages propagate through the Chain-of-Thought process, thereby amplifying the overall fragility of Visual CoT.

\begin{figure}[t]
    \centering
    \begin{minipage}[t]{0.46\textwidth}
        \centering
        \includegraphics[width=\linewidth]{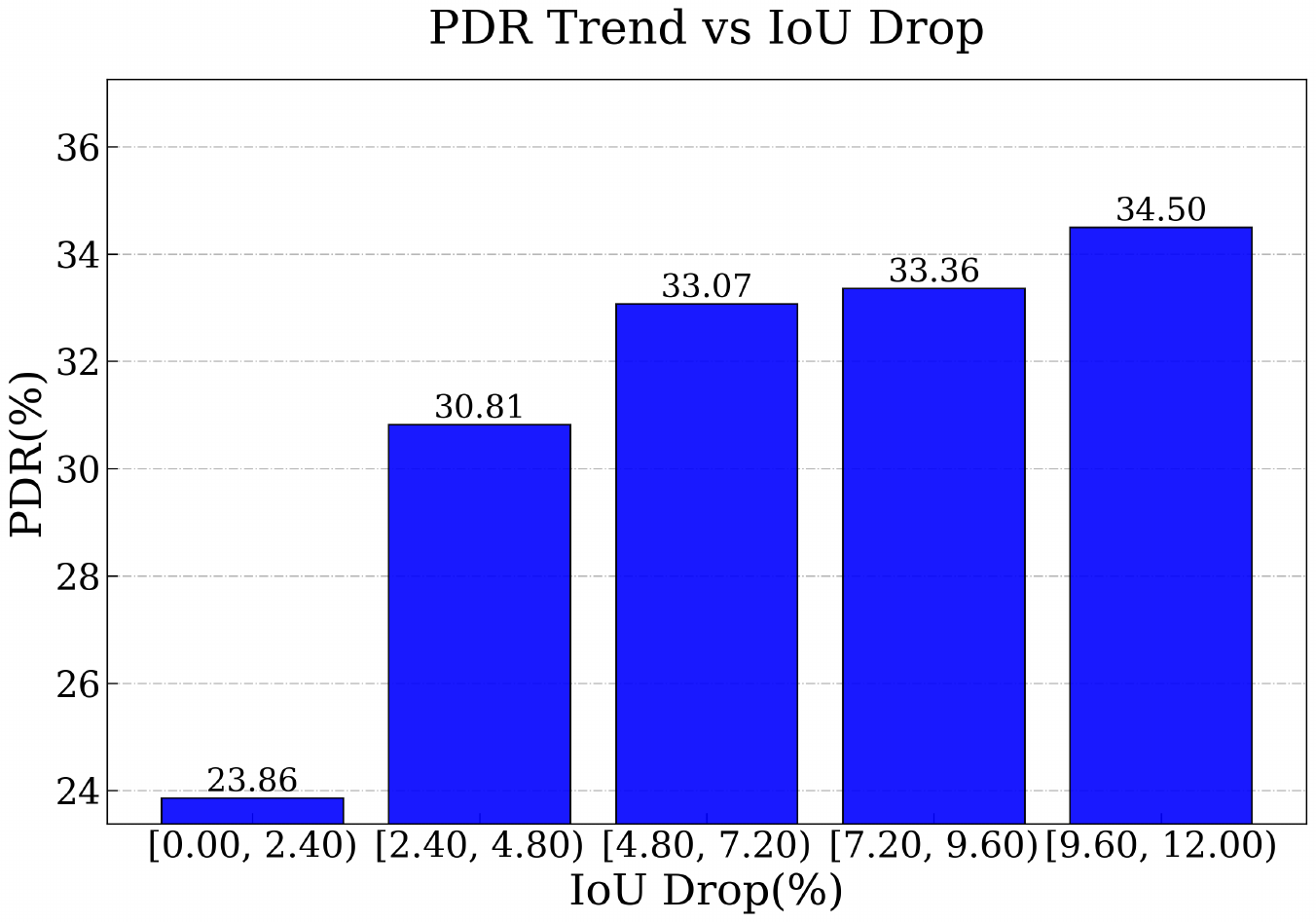}
        \caption{Correlation between PDR and IoU Degradation.
        }
        \label{fig:iou_drop}
    \end{minipage}
    \begin{minipage}[t]{0.48\textwidth}
        \centering
        \includegraphics[width=\linewidth]{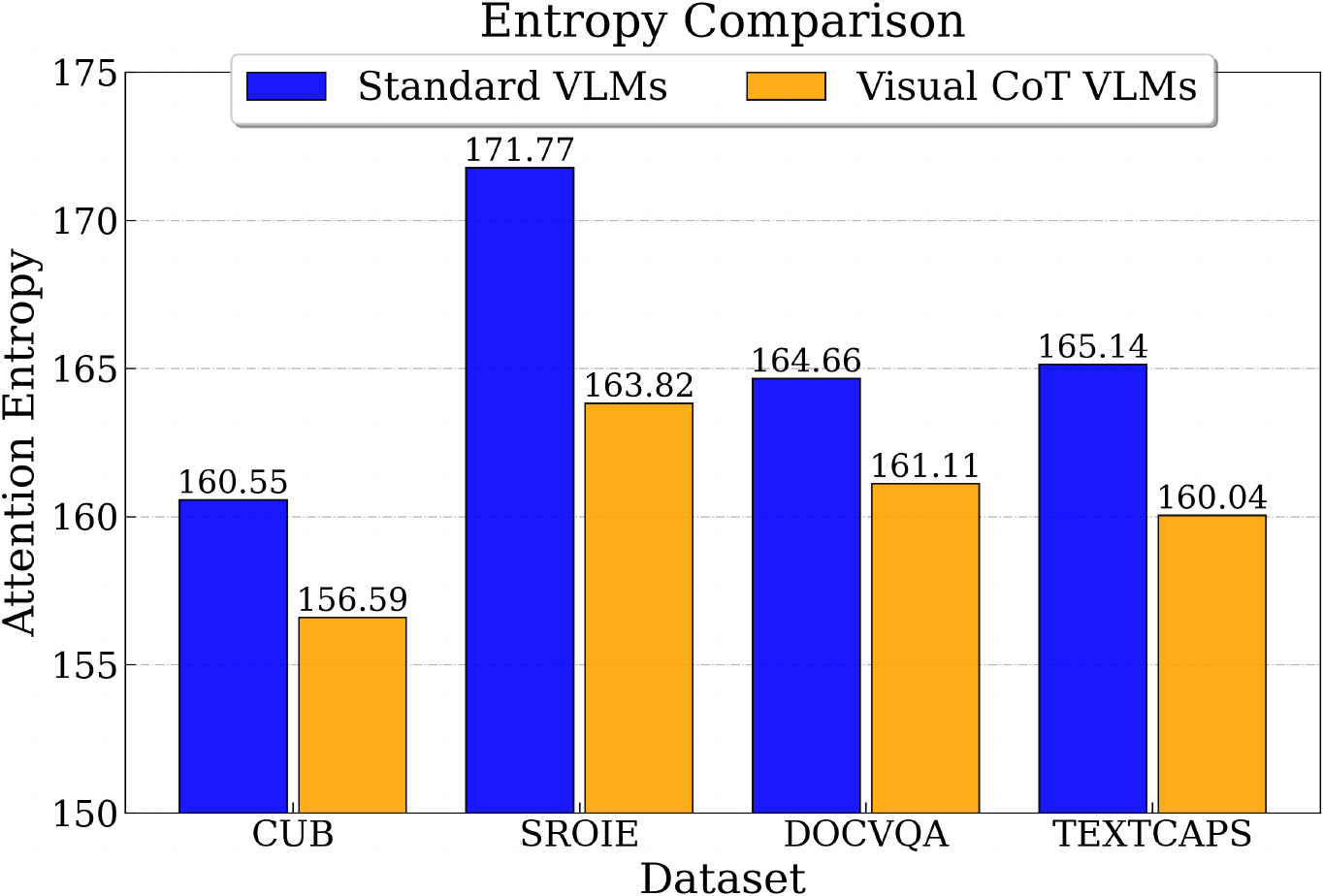}
        \caption{Average Attention Entropy.
        }
        \label{fig:attn_entropy}
    \end{minipage}
    \hfill
\end{figure}


\subsection{Can Attention Maps Explain Why \viscot Performs Better Under Perturbation?}

\begin{figure}
    \centering
    \includegraphics[width=1\linewidth]{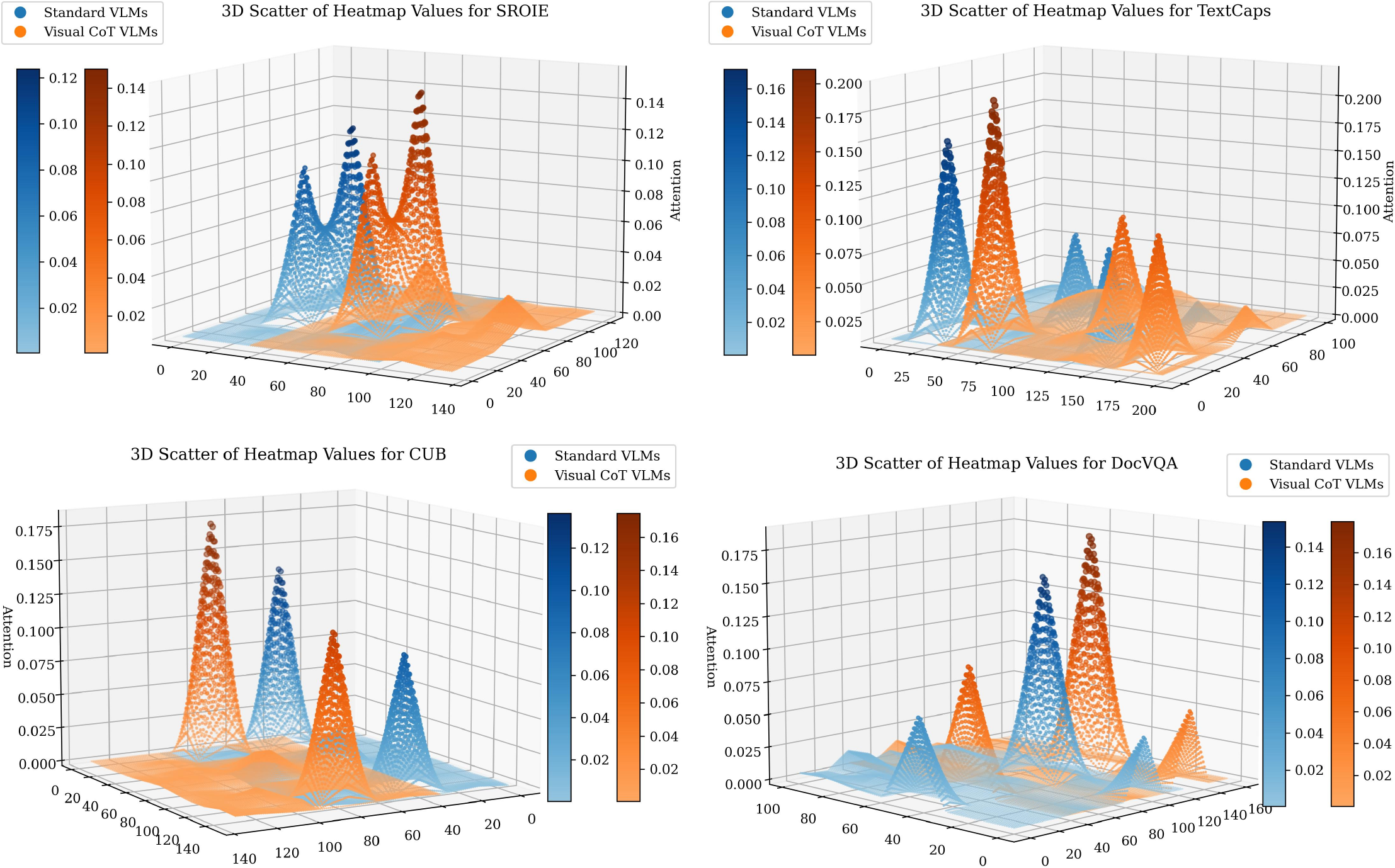}
    \caption{3D scatter plots of attention scores under noise conditions for Standard and Visual CoT VLMs.
Visual CoT exhibits more concentrated attention over key regions.}
    \label{fig:3d-attention}
\end{figure}

To better understand why \viscot achieve higher accuracy than \pvqa under noisy image conditions, we analyze the 3D attention distributions obtained from perturbed inputs, as shown in \Cref{fig:3d-attention}.
These visualizations reveal a clear distinction between the two paradigms:
while \pvqa tend to spread attention across broader regions with multiple dispersed peaks, \viscot produce more concentrated and sharper attention peaks focused on specific regions of the input.

This visual difference indicates that \viscot allocate their attention more selectively, focusing on semantically relevant areas while suppressing irrelevant regions. To quantitatively support this observation, we compute the attention entropy for each data instance. Specifically, a lower entropy value indicates that the model's attention is concentrated on more specific regions, whereas a higher entropy suggests a more dispersed focus. As shown in \Cref{fig:attn_entropy}, \viscot consistently exhibits lower entropy across samples, confirming a narrower and more focused attention distribution compared to \pvqa.

Such concentrated attention behavior helps \viscot better withstand noisy inputs by minimizing distractions from irrelevant tokens. In contrast, the broader and more uniform attention of \pvqa may dilute the impact of informative cues, reducing answer reliability under noise.


\begin{figure}[t]
    \centering
    \includegraphics[width=1\linewidth]{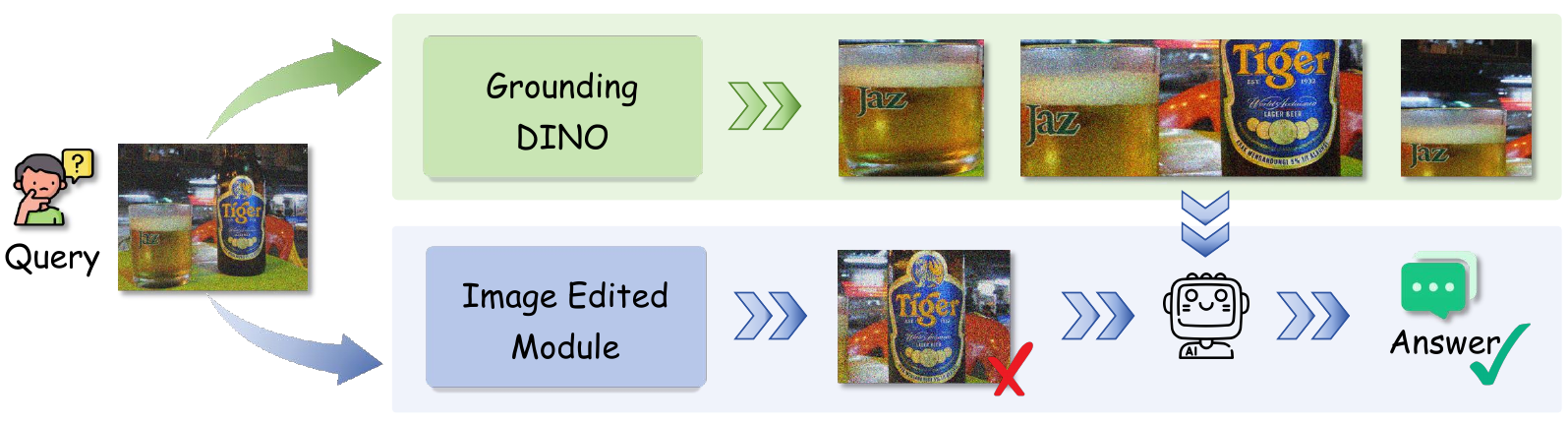}
    \caption{Enhanced Visual CoT pipeline with Grounding DINO.}
    \label{fig:gdino}
\end{figure}

\section{Robustness Enhancement}

Based on the above analysis, we argue that the multi-step reasoning process in \viscot acts as a double-edged sword: while it drives superior performance through structured reasoning and multimodal integration, it also introduces additional vulnerability to perturbations due to longer reasoning chains. Among these, the intermediate components, namely the local image patches play a pivotal role.
Consequently, strengthening the components
offers a promising and feasible direction to enhance \viscot’ robustness against noisy perturbations. In the following, we present a new approach to achieving this goal.


\subsection{Incorporating Grounding DINO for Enhanced Visual Information}


To enhance the robustness of \viscot under visual perturbations, we incorporate an auxiliary visual grounding step into the reasoning pipeline in a plug-and-play manner (as shown in \Cref{fig:gdino}). This step complements the original single-region strategy by identifying multiple semantically relevant regions that may provide redundant visual cues under noisy conditions.

Specifically, given an image–question pair, we apply Grounding DINO to generate text-conditioned region proposals. All bounding boxes with confidence scores exceeding a threshold (typically 0.4) are retained, and the selected regions are cropped from the original image. 
Then these auxiliary patches are appended to the original visual inputs and subsequently used within \viscot as supplementary visual cues to assist answer generation. 
This design encourages \vlm to attend to diverse visual perspectives during multi-step reasoning, thereby improving robustness under perturbations.


\begin{figure}[h]
    \centering

    \begin{subfigure}[t]{0.49\textwidth}
        \includegraphics[width=\linewidth]{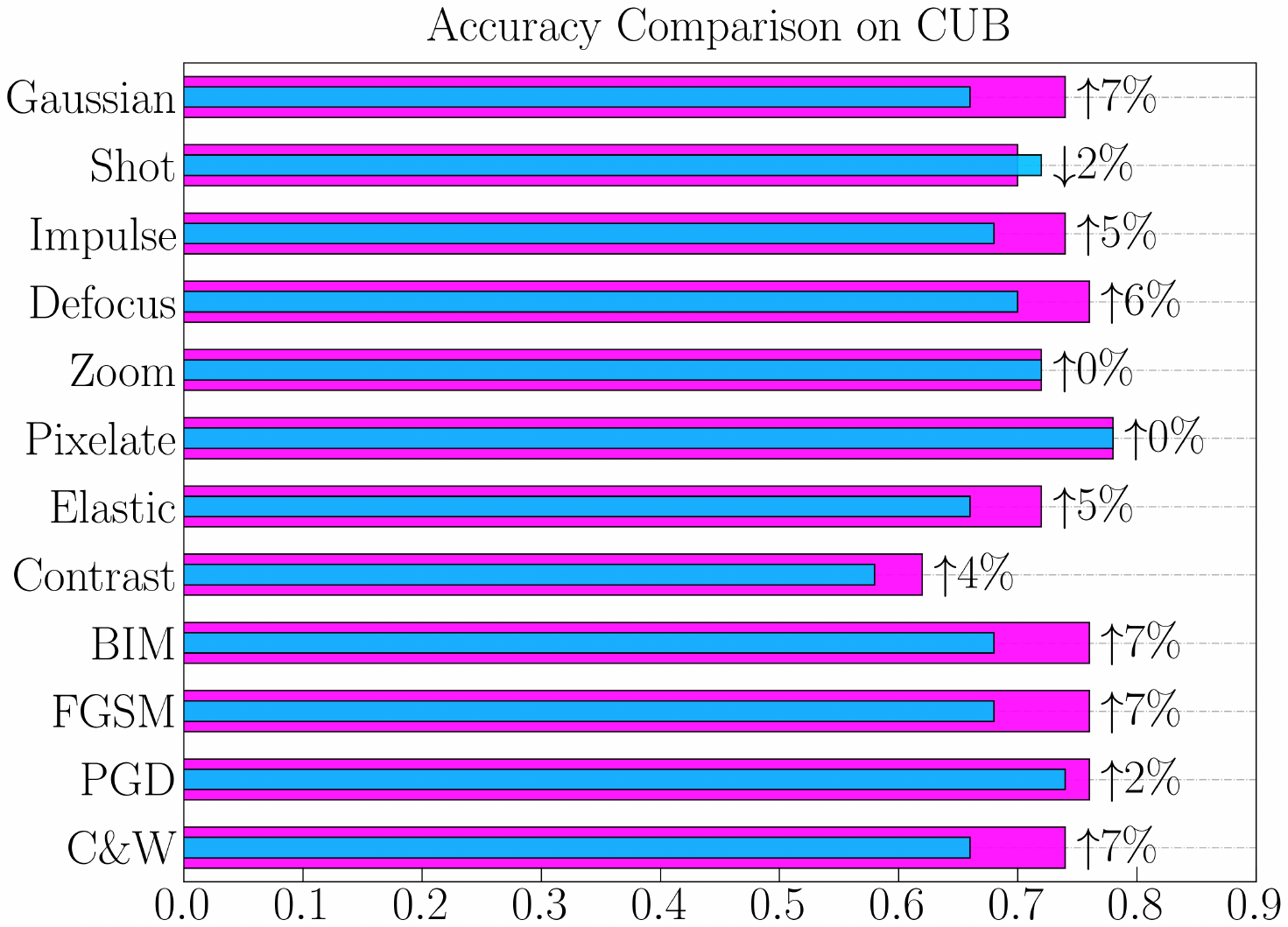}
    \end{subfigure}
    \hfill
    \begin{subfigure}[t]{0.49\textwidth}
        \includegraphics[width=\linewidth]{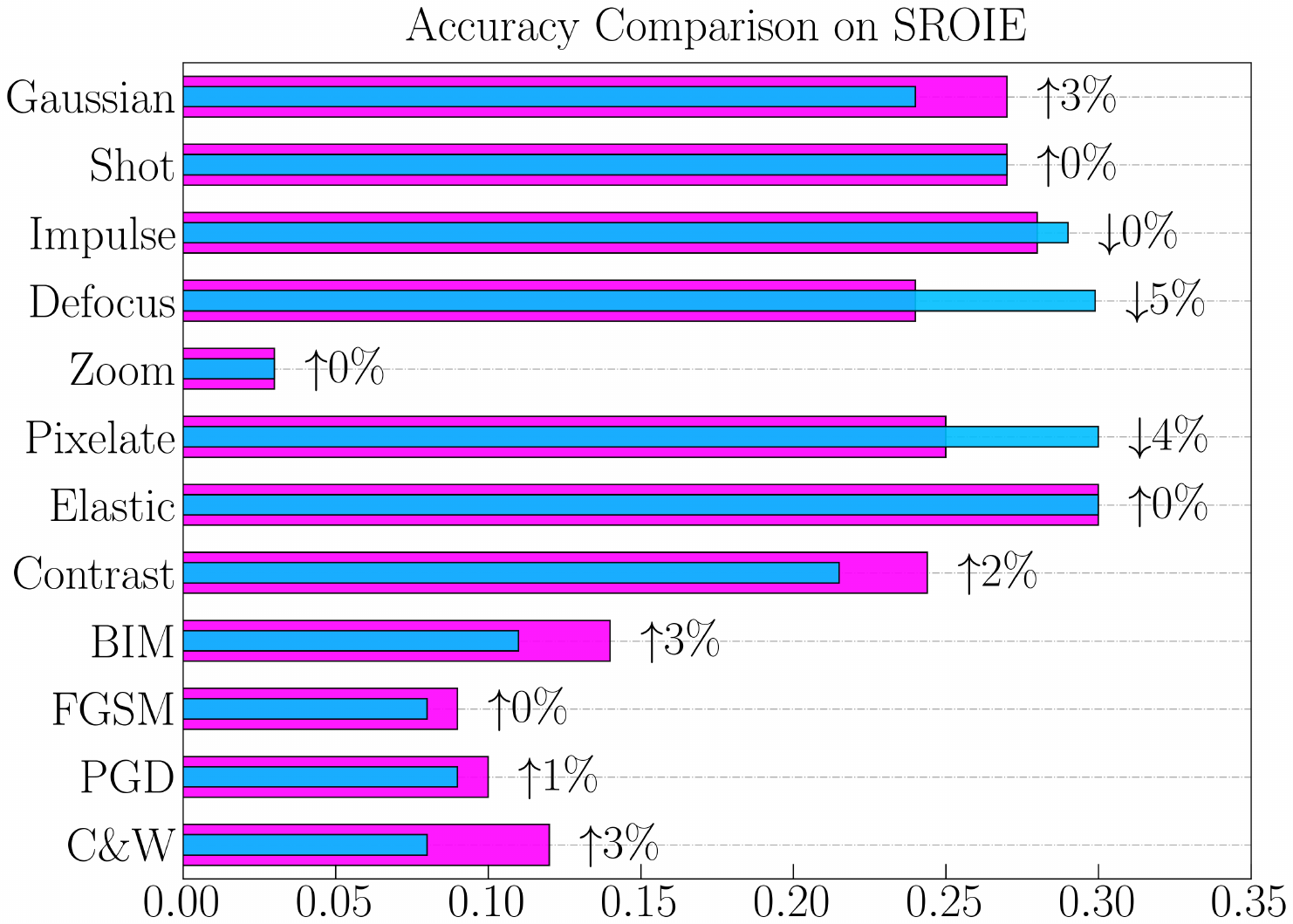}
    \end{subfigure}

    \vspace{0.1cm}

    \begin{subfigure}[t]{0.49\textwidth}
        \includegraphics[width=\linewidth]{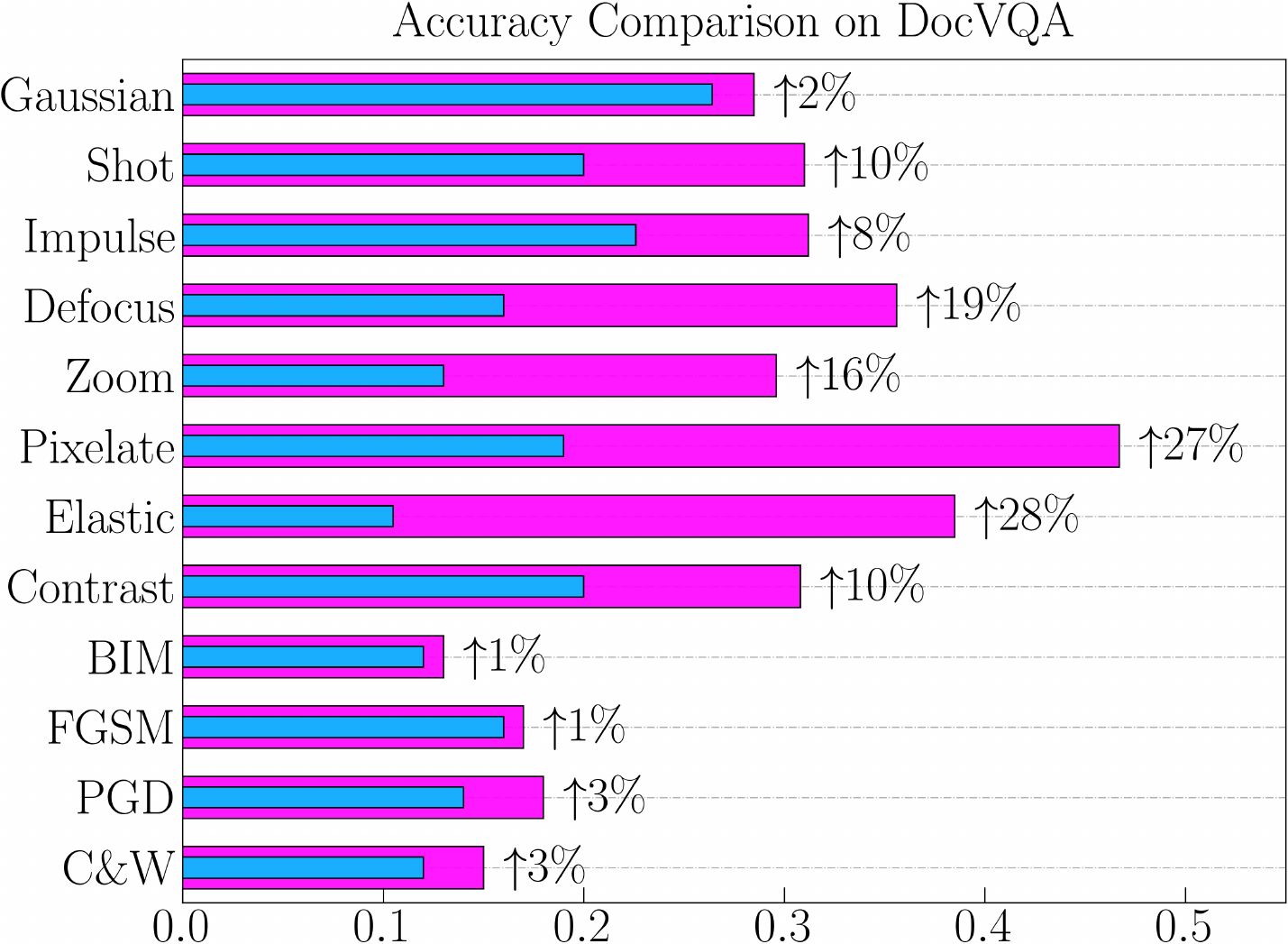}
    \end{subfigure}
    \hfill
    \begin{subfigure}[t]{0.49\textwidth}
        \includegraphics[width=\linewidth]{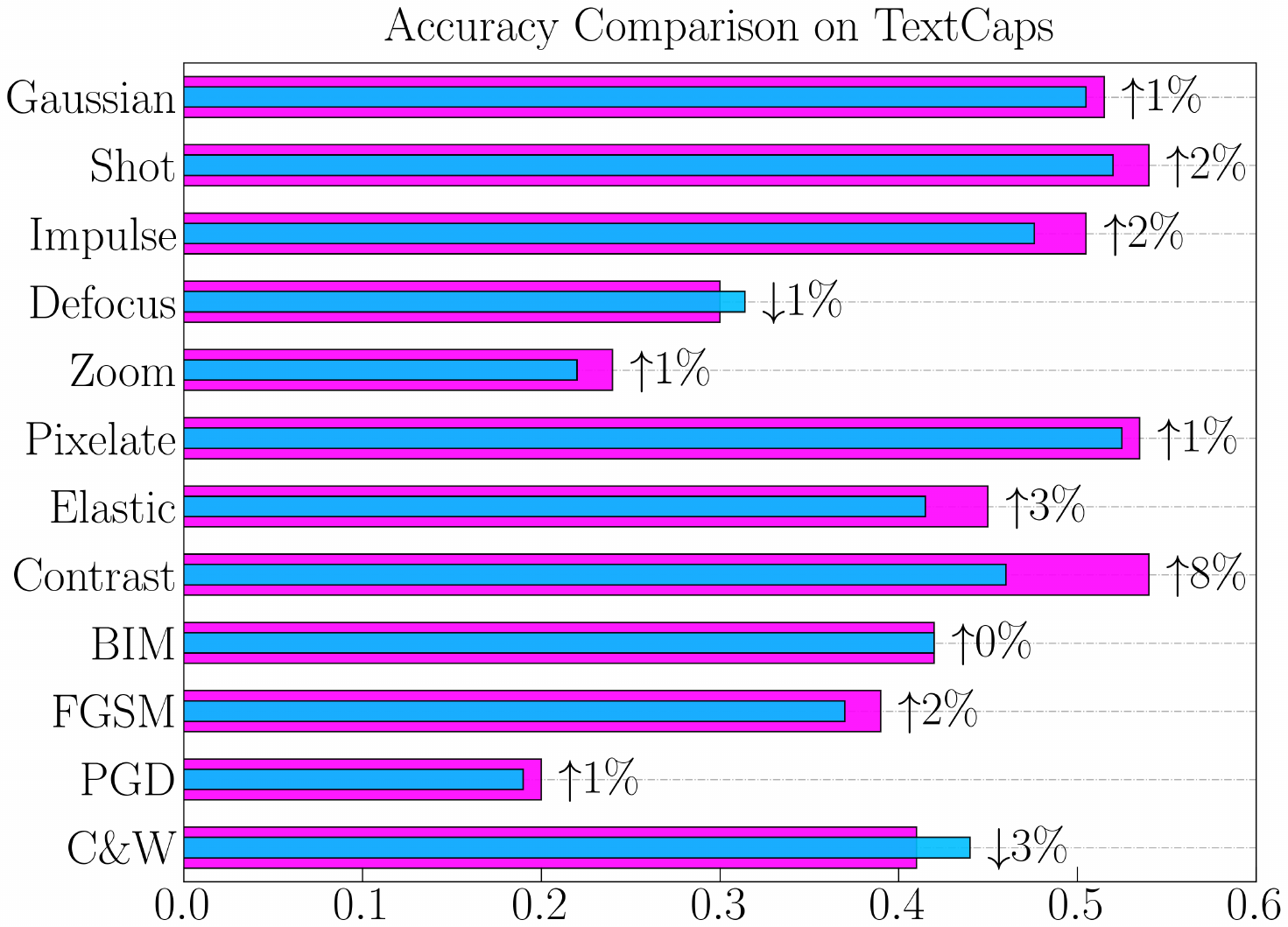}
    \end{subfigure}

    \vspace{0.1cm}
    \includegraphics[width=0.6\textwidth]{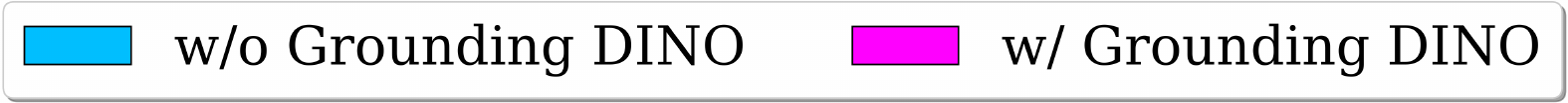}

    \caption{Accuracy comparison of Visual CoT with and without Grounding DINO under different image perturbations across four datasets.}
    \label{fig:gdino_result}
\end{figure}

\subsection{Experimental Results and Analysis}
Experimental results in \Cref{fig:gdino_result} demonstrate that, across all types of visual perturbations, the integration of Grounding DINO consistently enhances \viscot performance, yielding noticeable accuracy gains on the majority of evaluated datasets. On average, the incorporation of Grounding DINO leads to a 6\% increase in accuracy. The improvements are most pronounced on DocVQA, where accuracy gains frequently exceed 10\% under perturbations such as Pixelate, Elastic Transformations and Contrast Adjustments.

This improvement can be primarily attributed to Grounding DINO's ability to identify key regions in images. 
By integrating target bounding box information relevant to the given question into the Visual CoT reasoning process, the system's ability to resist noise interference and extract critical information is effectively strengthened. 
These findings suggest that incorporating visual grounding model into the reasoning process can significantly enhance the robustness of \viscot under perturbations.

\section{Conclusion}
In this paper, we present a systematic robustness study of Visual CoT reasoning in VLMs, revealing a fundamental trade-off: while Visual CoT improves answer accuracy, it also introduces increased sensitivity to visual perturbations. 
To address this limitation, we introduce a plug-and-play enhancement based on the Grounding DINO model, which improves robustness without requiring retraining of the base VLMs. Our work provides a foundation for future research on developing robust and reliable multimodal reasoning systems.

\bibliography{iclr2026_conference}

\begin{thebibliography}{26}
\providecommand{\natexlab}[1]{#1}
\providecommand{\url}[1]{\texttt{#1}}
\expandafter\ifx\csname urlstyle\endcsname\relax
  \providecommand{\doi}[1]{doi: #1}\else
  \providecommand{\doi}{doi: \begingroup \urlstyle{rm}\Url}\fi

\bibitem[Carlini \& Wagner(2016)Carlini and Wagner]{Carlini2016TowardsET}
Nicholas Carlini and David~A. Wagner.
\newblock Towards evaluating the robustness of neural networks.
\newblock \emph{2017 IEEE Symposium on Security and Privacy (SP)}, pp.\  39--57, 2016.
\newblock URL \url{https://api.semanticscholar.org/CorpusID:2893830}.

\bibitem[Chen et~al.(2024)Chen, Zhou, Shen, Hong, Sun, Gutfreund, and Gan]{Chen_Zhou_Shen_2024}
Zhenfang Chen, Qinhong Zhou, Yikang Shen, Yining Hong, Zhiqing Sun, Dan Gutfreund, and Chuang Gan.
\newblock Visual chain-of-thought prompting for knowledge-based visual reasoning.
\newblock \emph{Proceedings of the AAAI Conference on Artificial Intelligence}, 38\penalty0 (2):\penalty0 1254--1262, Mar. 2024.
\newblock \doi{10.1609/aaai.v38i2.27888}.
\newblock URL \url{https://ojs.aaai.org/index.php/AAAI/article/view/27888}.

\bibitem[Fu et~al.(2025)Fu, Liu, Yang, Corring, Lu, Yang, Roth, Flor{\^e}ncio, and Zhang]{Fu2025ReFocusVE}
Xingyu Fu, Minqian Liu, Zhengyuan Yang, John Corring, Yijuan Lu, Jianwei Yang, Dan Roth, Dinei A.~F. Flor{\^e}ncio, and Cha Zhang.
\newblock Refocus: Visual editing as a chain of thought for structured image understanding.
\newblock \emph{ArXiv}, abs/2501.05452, 2025.
\newblock URL \url{https://api.semanticscholar.org/CorpusID:275405594}.

\bibitem[Gao et~al.(2024)Gao, Bai, Bai, Yang, and Xia]{Gao2024AdversarialRF}
Kuofeng Gao, Yang Bai, Jiawang Bai, Yong Yang, and Shu-Tao Xia.
\newblock Adversarial robustness for visual grounding of multimodal large language models.
\newblock \emph{ArXiv}, abs/2405.09981, 2024.
\newblock URL \url{https://api.semanticscholar.org/CorpusID:269791017}.

\bibitem[Goodfellow et~al.(2014)Goodfellow, Shlens, and Szegedy]{Goodfellow2014ExplainingAH}
Ian~J. Goodfellow, Jonathon Shlens, and Christian Szegedy.
\newblock Explaining and harnessing adversarial examples.
\newblock \emph{CoRR}, abs/1412.6572, 2014.
\newblock URL \url{https://api.semanticscholar.org/CorpusID:6706414}.

\bibitem[Hendrycks \& Dietterich(2019)Hendrycks and Dietterich]{HendrycksD19}
Dan Hendrycks and Thomas~G. Dietterich.
\newblock Benchmarking neural network robustness to common corruptions and perturbations.
\newblock In \emph{7th International Conference on Learning Representations, {ICLR} 2019, New Orleans, LA, USA, May 6-9, 2019}. OpenReview.net, 2019.
\newblock URL \url{https://openreview.net/forum?id=HJz6tiCqYm}.

\bibitem[Hu et~al.(2024)Hu, Shi, Fu, Roth, Ostendorf, Zettlemoyer, Smith, and Krishna]{hu2024visual}
Yushi Hu, Weijia Shi, Xingyu Fu, Dan Roth, Mari Ostendorf, Luke Zettlemoyer, Noah~A Smith, and Ranjay Krishna.
\newblock Visual sketchpad: Sketching as a visual chain of thought for multimodal language models.
\newblock \emph{Advances in Neural Information Processing Systems}, 37:\penalty0 139348--139379, 2024.

\bibitem[Huang et~al.(2019)Huang, Chen, He, Bai, Karatzas, Lu, and Jawahar]{DBLP:conf/icdar/HuangCHBKLJ19}
Zheng Huang, Kai Chen, Jianhua He, Xiang Bai, Dimosthenis Karatzas, Shijian Lu, and C.~V. Jawahar.
\newblock {ICDAR2019} competition on scanned receipt {OCR} and information extraction.
\newblock In \emph{2019 International Conference on Document Analysis and Recognition, {ICDAR} 2019, Sydney, Australia, September 20-25, 2019}, pp.\  1516--1520. {IEEE}, 2019.
\newblock \doi{10.1109/ICDAR.2019.00244}.
\newblock URL \url{https://doi.org/10.1109/ICDAR.2019.00244}.

\bibitem[Jiang et~al.(2025{\natexlab{a}})Jiang, Zhang, Guo, Li, Qi, Chen, Wang, Jin, Guo, Yan, et~al.]{jiang2025mme}
Dongzhi Jiang, Renrui Zhang, Ziyu Guo, Yanwei Li, Yu~Qi, Xinyan Chen, Liuhui Wang, Jianhan Jin, Claire Guo, Shen Yan, et~al.
\newblock Mme-cot: Benchmarking chain-of-thought in large multimodal models for reasoning quality, robustness, and efficiency.
\newblock \emph{arXiv preprint arXiv:2502.09621}, 2025{\natexlab{a}}.

\bibitem[Jiang et~al.(2025{\natexlab{b}})Jiang, Xu, Singh, and Singh]{jiang2025misaligning}
Enyi Jiang, Changming Xu, Nischay Singh, and Gagandeep Singh.
\newblock Misaligning reasoning with answers--a framework for assessing llm cot robustness.
\newblock \emph{arXiv preprint arXiv:2505.17406}, 2025{\natexlab{b}}.

\bibitem[Jin et~al.(2024)Jin, Li, Guo, Su, Zhang, and Zeng]{jin2024saber}
Naizhu Jin, Zhong Li, Yinggang Guo, Chao Su, Tian Zhang, and Qingkai Zeng.
\newblock Saber: Model-agnostic backdoor attack on chain-of-thought in neural code generation.
\newblock \emph{arXiv preprint arXiv:2412.05829}, 2024.

\bibitem[Kurakin et~al.(2017)Kurakin, Goodfellow, and Bengio]{kurakin2017adversarial}
Alexey Kurakin, Ian Goodfellow, and Samy Bengio.
\newblock Adversarial examples in the physical world.
\newblock In \emph{Proceedings of the 5th International Conference on Learning Representations (ICLR) Workshop}, 2017.

\bibitem[Liu et~al.(2023{\natexlab{a}})Liu, Li, Wu, and Lee]{liu2023llava}
Haotian Liu, Chunyuan Li, Qingyang Wu, and Yong~Jae Lee.
\newblock Visual instruction tuning, 2023{\natexlab{a}}.

\bibitem[Liu et~al.(2023{\natexlab{b}})Liu, Zeng, Ren, Li, Zhang, Yang, Li, Yang, Su, Zhu, et~al.]{liu2023grounding}
Shilong Liu, Zhaoyang Zeng, Tianhe Ren, Feng Li, Hao Zhang, Jie Yang, Chunyuan Li, Jianwei Yang, Hang Su, Jun Zhu, et~al.
\newblock Grounding dino: Marrying dino with grounded pre-training for open-set object detection.
\newblock \emph{arXiv preprint arXiv:2303.05499}, 2023{\natexlab{b}}.

\bibitem[Makelov et~al.(2025)Makelov, Lange, and Nanda]{DBLP:conf/iclr/MakelovLN25}
Aleksandar Makelov, Georg Lange, and Neel Nanda.
\newblock Towards principled evaluations of sparse autoencoders for interpretability and control.
\newblock In \emph{The Thirteenth International Conference on Learning Representations, {ICLR} 2025, Singapore, April 24-28, 2025}. OpenReview.net, 2025.

\bibitem[Mathew et~al.(2021)Mathew, Karatzas, and Jawahar]{mathew2021docvqa}
Minesh Mathew, Dimosthenis Karatzas, and CV~Jawahar.
\newblock Docvqa: A dataset for vqa on document images.
\newblock In \emph{Proceedings of the IEEE/CVF winter conference on applications of computer vision}, pp.\  2200--2209, 2021.

\bibitem[Shao et~al.(2024)Shao, Qian, Xiao, Song, Zong, Wang, Liu, and Li]{shao2024visual}
Hao Shao, Shengju Qian, Han Xiao, Guanglu Song, Zhuofan Zong, Letian Wang, Yu~Liu, and Hongsheng Li.
\newblock Visual cot: Advancing multi-modal language models with a comprehensive dataset and benchmark for chain-of-thought reasoning.
\newblock \emph{Advances in Neural Information Processing Systems}, 37:\penalty0 8612--8642, 2024.

\bibitem[Sidorov et~al.(2020)Sidorov, Hu, Rohrbach, and Singh]{sidorov2020textcaps}
Oleksii Sidorov, Ronghang Hu, Marcus Rohrbach, and Amanpreet Singh.
\newblock Textcaps: a dataset for image captioning with reading comprehension.
\newblock In \emph{European conference on computer vision}, pp.\  742--758. Springer, 2020.

\bibitem[Wah et~al.(2011)Wah, Branson, Welinder, Perona, and Belongie]{Wah2011TheCB}
Catherine Wah, Steve Branson, Peter Welinder, Pietro Perona, and Serge~J. Belongie.
\newblock The caltech-ucsd birds-200-2011 dataset.
\newblock 2011.
\newblock URL \url{https://api.semanticscholar.org/CorpusID:16119123}.

\bibitem[Wang et~al.(2025)Wang, Wang, Cheng, Fei, Ding, Guo, Tao, and Qiu]{wang2025visuothink}
Yikun Wang, Siyin Wang, Qinyuan Cheng, Zhaoye Fei, Liang Ding, Qipeng Guo, Dacheng Tao, and Xipeng Qiu.
\newblock Visuothink: Empowering lvlm reasoning with multimodal tree search.
\newblock \emph{arXiv preprint arXiv:2504.09130}, 2025.

\bibitem[Wang et~al.(2024)Wang, Han, Chen, Xue, Ding, Xiao, Tresp, Torr, and Gu]{wang2024stop}
Zefeng Wang, Zhen Han, Shuo Chen, Fan Xue, Zifeng Ding, Xun Xiao, Volker Tresp, Philip Torr, and Jindong Gu.
\newblock Stop reasoning! when multimodal llm with chain-of-thought reasoning meets adversarial image.
\newblock \emph{arXiv preprint arXiv:2402.14899}, 2024.

\bibitem[Yang et~al.(2023)Yang, Li, Wang, Lin, Azarnasab, Ahmed, Liu, Liu, Zeng, and Wang]{yang2023mm}
Zhengyuan Yang, Linjie Li, Jianfeng Wang, Kevin Lin, Ehsan Azarnasab, Faisal Ahmed, Zicheng Liu, Ce~Liu, Michael Zeng, and Lijuan Wang.
\newblock Mm-react: Prompting chatgpt for multimodal reasoning and action.
\newblock \emph{arXiv preprint arXiv:2303.11381}, 2023.

\bibitem[Zhang et~al.(2024)Zhang, Yang, Lyu, Jin, Yao, Chen, and Luo]{zhang2024cocot}
Daoan Zhang, Junming Yang, Hanjia Lyu, Zijian Jin, Yuan Yao, Mingkai Chen, and Jiebo Luo.
\newblock Cocot: Contrastive chain-of-thought prompting for large multimodal models with multiple image inputs.
\newblock \emph{arXiv preprint arXiv:2401.02582}, 2024.

\bibitem[Zhou et~al.(2024{\natexlab{a}})Zhou, Tao, Zhu, Luo, Wang, and Han]{zhou2024can}
Zhanke Zhou, Rong Tao, Jianing Zhu, Yiwen Luo, Zengmao Wang, and Bo~Han.
\newblock Can language models perform robust reasoning in chain-of-thought prompting with noisy rationales?
\newblock \emph{Advances in Neural Information Processing Systems}, 37:\penalty0 123846--123910, 2024{\natexlab{a}}.

\bibitem[Zhou et~al.(2024{\natexlab{b}})Zhou, Tao, Zhu, Luo, Wang, and Han]{zhou_can_2024}
Zhanke Zhou, Rong Tao, Jianing Zhu, Yiwen Luo, Zengmao Wang, and Bo~Han.
\newblock Can language models perform robust reasoning in chain-of-thought prompting with noisy rationales?
\newblock In A.~Globerson, L.~Mackey, D.~Belgrave, A.~Fan, U.~Paquet, J.~Tomczak, and C.~Zhang (eds.), \emph{Advances in Neural Information Processing Systems}, volume~37, pp.\  123846--123910. Curran Associates, Inc., 2024{\natexlab{b}}.

\bibitem[Zhu et~al.(2024)Zhu, Wang, Zhou, Wang, Chen, Wang, Yang, Ye, Zhang, Gong, and Xie]{zhu2024promptrobustevaluatingrobustnesslarge}
Kaijie Zhu, Jindong Wang, Jiaheng Zhou, Zichen Wang, Hao Chen, Yidong Wang, Linyi Yang, Wei Ye, Yue Zhang, Neil~Zhenqiang Gong, and Xing Xie.
\newblock Promptrobust: Towards evaluating the robustness of large language models on adversarial prompts, 2024.
\newblock URL \url{https://arxiv.org/abs/2306.04528}.

\end{thebibliography}
\bibliographystyle{iclr2026_conference}

\newpage
\appendix
\section{Appendix}

\subsection{Use of Large Language Models}

To improve the readability of this manuscript, we used LLMs for language polishing, such as rephrasing sentences for clarity and correcting grammar. The LLMs were not involved in designing the methodology, conducting experiments, analyzing results, or drawing scientific conclusions. All substantive research contributions are the sole work of the authors.

\subsection{Severity Levels of Natural Perturbations
}
\label{severity_natural}
For all natural corruption methods, we adopt five severity levels from 1 to 5, following the ImageNet-C benchmark design. Severity controls the intensity of distortion, with level 1 corresponding to the weakest corruption and level 5 to the strongest. Below we explain the parameterization of each corruption and how these parameters influence the degree of degradation.

\begin{table*}[h]
\centering
\renewcommand{\arraystretch}{1.2}
\caption{Severity Level Settings for Visual Perturbations (Code-based Implementation). Severity increases from 1 to 5, with higher levels indicating stronger perturbations.}
\label{tab:vis_perturbation_severity_code}
\resizebox{\textwidth}{!}{
\begin{tabular}{c|c|c|p{6.5cm}}
\toprule
\textbf{Perturbation} & \textbf{Parameter} & \textbf{Severity Values (1→5)} & \textbf{Effect Description} \\
\midrule

Gaussian Noise & Std. dev. $\sigma$ & [0.08, 0.12, 0.18, 0.26, 0.38] & Adds Gaussian-distributed pixel noise. Larger $\sigma$ → stronger random fluctuations, fine details vanish at severity 5. \\
\midrule

Shot Noise & Photon count scale $c$ & [60, 25, 12, 5, 3] & Lower $c$ → stronger Poisson noise. Severity 5 simulates extreme low-light, heavy discrete fluctuations. \\
\midrule

Impulse Noise & Pixel corruption ratio $p$ & [0.03, 0.06, 0.09, 0.17, 0.27] & Higher $p$ replaces more pixels with black/white. Severity 5 → $\sim$27\% pixels corrupted. \\
\midrule

Defocus Blur & Disk radius, alias blur & [(3,0.1), (4,0.5), (6,0.5), (8,0.5), (10,0.5)] & Increasing radius → stronger out-of-focus blur. At severity 5, edges/boundaries disappear. \\
\midrule

Zoom Blur & Zoom factor ranges $c$ & 
\begin{tabular}[c]{@{}l@{}} 
1: 1.00–1.10 (step 0.01) \\ 
2: 1.00–1.15 (step 0.01) \\ 
3: 1.00–1.21 (step 0.02) \\ 
4: 1.00–1.26 (step 0.02) \\ 
5: 1.00–1.33 (step 0.03) 
\end{tabular} 
& Combines zoomed-in frames. Larger ranges represent stronger radial streaks. Severity 5 represents heavy smearing. \\
\midrule

Pixelate & Downsample ratio $c$ & [0.6, 0.5, 0.4, 0.3, 0.25] & Image is resized to $c \cdot$ original then upscaled. Lowe  r values → larger blocks. Severity 5 → coarse blockiness. \\
\midrule

Elastic Transform & $(\alpha, \sigma, \alpha_{affine})$ & 
\begin{tabular}[c]{@{}l@{}} 
1: (224$\times$2, 224$\times$0.7, 224$\times$0.1) \\ 
2: (224$\times$2, 224$\times$0.08, 224$\times$0.2) \\ 
3: (224$\times$0.05, 224$\times$0.01, 224$\times$0.02) \\ 
4: (224$\times$0.07, 224$\times$0.01, 224$\times$0.02) \\ 
5: (224$\times$0.12, 224$\times$0.01, 224$\times$0.02) 
\end{tabular} 
& $\alpha$ controls displacement, $\sigma$ smooths deformation, $\alpha_{affine}$ adds affine distortion. Higher severity → stronger warping, shapes deformed. \\
\midrule

Contrast Reduction & Contrast scaling $c$ & [0.4, 0.3, 0.2, 0.1, 0.05] & Lower $c$ reduces pixel variance. Severity 5 → image looks flat/washed out. \\
\bottomrule

\end{tabular}
}
\end{table*}

\subsection{Severity Levels of White-box Adversarial Attack}
\label{severity_White_box}

The four standard adversarial attacks: FGSM, BIM, PGD, and C\&W, are conducted on the image encoder of the model (e.g., CLIP ViT or other vision towers), targeting the embedding similarity between clean and adversarial samples. The goal is to cause minimal pixel changes while maximally deviating the internal representations \cite{Gao2024AdversarialRF}.

\subsubsection{Fast Gradient Sign Method (FGSM)}
\begin{algorithm}[H]
\caption{FGSM Algorithm}
\begin{algorithmic}[1]
\Require Input image $x$, vision encoder $f(\cdot)$, loss function $\mathcal{L}$ (e.g., MSE), perturbation bound $\epsilon$
\Ensure Adversarial image $x_{\text{adv}}$
\State Compute clean embedding: $z_{\text{clean}} \gets f(x)$
\State Initialize perturbation: $\delta \gets 0$, set $\delta$ as a trainable tensor
\State $z_{\text{adv}} \gets f(x + \delta)$
\State $\mathcal{L}_{\text{adv}} \gets \text{MSE}(z_{\text{adv}}, z_{\text{clean}})$
\State Compute gradient: $\nabla_{\delta} \mathcal{L}_{\text{adv}}$
\State Update perturbation: $\delta \gets \epsilon \cdot \text{sign}(\nabla_{\delta} \mathcal{L}_{\text{adv}})$
\State $x_{\text{adv}} \gets \text{clip}(x + \delta, 0, 1)$ \Comment{Ensure valid pixel range}
\State \Return $x_{\text{adv}}$
\end{algorithmic}
\end{algorithm}

\begin{table}[H]
\centering
\caption{Severity Level Settings for FGSM Attack}
\label{tab:fgsm_severity}
\renewcommand{\arraystretch}{1.2}
\begin{tabular}{c|c}
\toprule
\textbf{Severity Level} & $\boldsymbol{\epsilon}$ (Perturbation Magnitude) \\
\midrule
1 & $\frac{1}{255}$ \\
2 & $\frac{2}{255}$ \\
3 & $\frac{4}{255}$ \\
4 & $\frac{6}{255}$ \\
5 & $\frac{8}{255}$ \\
\bottomrule
\end{tabular}
\end{table}

\subsubsection{Basic Iterative Method (BIM)}

\begin{algorithm}[H]
\caption{BIM Algorithm}
\begin{algorithmic}[1]
\Require Input image $x$, vision encoder $f(\cdot)$, loss function $\mathcal{L}$ (e.g., MSE), step size $\alpha$, maximum perturbation $\epsilon$, number of iterations $T$
\Ensure Adversarial image $x_{\text{adv}}$
\State Compute clean embedding $z_{\text{clean}} \gets f(x)$
\State Initialize perturbation $\delta \gets \mathbf{0}$ \Comment{No random start}
\For{$t = 1$ to $T$}
    \State $z_{\text{adv}} \gets f(x + \delta)$ \Comment{Get perturbed embedding}
    \State $\mathcal{L}_{\text{adv}} \gets \text{MSE}(z_{\text{adv}}, z_{\text{clean}})$
    \State Compute gradient $\nabla_{\delta} \mathcal{L}_{\text{adv}}$
    \State $\delta \gets \delta + \alpha \cdot \text{sign}(\nabla_{\delta} \mathcal{L}_{\text{adv}})$
    \State $\delta \gets \text{clip}(\delta, -\epsilon, \epsilon)$ \Comment{Clip to $\ell_\infty$ ball}
    \State $\delta \gets \text{clip}(x + \delta, 0, 1) - x$ \Comment{Ensure pixel validity}
\EndFor
\State $x_{\text{adv}} \gets \text{clip}(x + \delta, 0, 1)$
\State \Return $x_{\text{adv}}$
\end{algorithmic}
\end{algorithm}

\begin{table}[H]
\centering
\caption{Severity Level Settings for BIM Attack}
\label{tab:bim_severity}
\renewcommand{\arraystretch}{1.2}
\begin{tabular}{c|c|c|c}
\toprule
\textbf{Severity Level} & $\boldsymbol{\epsilon}$ & $\boldsymbol{\alpha}$ (Step Size) & $\boldsymbol{T}$ (Iterations) \\
\midrule
1 & $\frac{1}{255}$   & $\frac{0.2}{255}$ & 100 \\
2 & $\frac{2}{255}$   & $\frac{0.4}{255}$ & 200 \\
3 & $\frac{4}{255}$   & $\frac{0.8}{255}$ & 300 \\
4 & $\frac{6}{255}$   & $\frac{1.0}{255}$ & 400 \\
5 & $\frac{8}{255}$   & $\frac{1.2}{255}$ & 500 \\
\bottomrule
\end{tabular}
\end{table}

\subsubsection{Projected Gradient Descent (PGD)}

\begin{algorithm}[H]
\caption{PGD Algorithm}
\begin{algorithmic}[1]
\Require Input image $x$, vision encoder $f(\cdot)$, loss function $\mathcal{L}$ (e.g., MSE), step size $\alpha$, maximum perturbation $\epsilon$, number of iterations $T$
\Ensure Adversarial image $x_{\text{adv}}$
\State Initialize perturbation $\delta \sim \text{Uniform}(-\epsilon, \epsilon)$
\For{$t = 1$ to $T$}
    \State $z_{\text{adv}} \gets f(x + \delta)$ \Comment{Get perturbed embedding}
    \State $z_{\text{clean}} \gets f(x)$ \Comment{(Optional) Use precomputed clean embedding}
    \State $\mathcal{L}_{\text{adv}} \gets \text{MSE}(z_{\text{adv}}, z_{\text{clean}})$
    \State Compute gradient $\nabla_{\delta} \mathcal{L}_{\text{adv}}$
    \State $\delta \gets \delta + \alpha \cdot \text{sign}(\nabla_{\delta} \mathcal{L}_{\text{adv}})$
    \State $\delta \gets \text{clip}(\delta, -\epsilon, \epsilon)$ \Comment{Project onto $\ell_\infty$ ball}
\EndFor
\State $x_{\text{adv}} \gets \text{clip}(x + \delta, 0, 1)$ \Comment{Clamp to valid pixel range}
\State \Return $x_{\text{adv}}$
\end{algorithmic}
\end{algorithm}

\begin{table}[H]
\centering
\caption{Severity Level Settings for PGD Attack}
\label{tab:pgd_severity}
\renewcommand{\arraystretch}{1.2}
\begin{tabular}{c|c|c|c}
\toprule
\textbf{Severity Level} & $\boldsymbol{\epsilon}$ & $\boldsymbol{\alpha}$ (Step Size) & $\boldsymbol{T}$ (Iterations) \\
\midrule
1 & $\frac{1}{255}$   & $\frac{0.2}{255}$ & 100 \\
2 & $\frac{2}{255}$   & $\frac{0.4}{255}$ & 200 \\
3 & $\frac{4}{255}$   & $\frac{0.8}{255}$ & 300 \\
4 & $\frac{6}{255}$   & $\frac{1.0}{255}$ & 400 \\
5 & $\frac{8}{255}$   & $\frac{1.2}{255}$ & 500 \\
\bottomrule
\end{tabular}
\end{table}

\subsubsection{Carlini \& Wagner (C\&W) Attack (Untargeted)}

\begin{algorithm}[H]
\caption{C\&W Attack Algorithm}
\begin{algorithmic}[1]
\Require Input image $x$, vision encoder $f(\cdot)$, loss function $\mathcal{L}$ (e.g., MSE), regularization coefficient $C$, learning rate $\eta$, number of iterations $T$
\Ensure Adversarial image $x_{\text{adv}}$
\State Compute clean embedding: $z_{\text{clean}} \gets f(x)$
\State Convert $x$ to tanh-space: $w \gets \text{arctanh}(2x - 1)$ \Comment{Ensure differentiability}
\For{$t = 1$ to $T$}
    \State $x_{\text{adv}} \gets 0.5 \cdot (\tanh(w) + 1)$ \Comment{Map $w$ back to $[0,1]$}
    \State $z_{\text{adv}} \gets f(x_{\text{adv}})$
    \State Compute adversarial loss: $\mathcal{L}_{\text{embed}} \gets \text{MSE}(z_{\text{adv}}, z_{\text{clean}})$
    \State Compute distortion loss: $\mathcal{L}_{\text{l2}} \gets \|x_{\text{adv}} - x\|_2^2$
    \State Total loss: $\mathcal{L}_{\text{total}} \gets C \cdot \mathcal{L}_{\text{embed}} + \mathcal{L}_{\text{l2}}$
    \State Update $w$ via Adam: $w \gets w - \eta \cdot \nabla_w \mathcal{L}_{\text{total}}$
\EndFor
\State \Return $x_{\text{adv}}$
\end{algorithmic}
\end{algorithm}

\begin{table}[H]
\centering
\caption{Severity Level Settings for C\&W Attack}
\label{tab:cw_severity}
\renewcommand{\arraystretch}{1.2}
\begin{tabular}{c|c|c|c}
\toprule
\textbf{Severity Level} & $\boldsymbol{C}$ (Embed Weight) & $\boldsymbol{\eta}$ (Learning Rate) & $\boldsymbol{T}$ (Iterations) \\
\midrule
1 & 0.1  & $1 \times 10^{-3}$ & 100 \\
2 & 0.5  & $1 \times 10^{-3}$ & 200 \\
3 & 1.0  & $5 \times 10^{-4}$ & 300 \\
4 & 2.0  & $1 \times 10^{-4}$ & 400 \\
5 & 5.0  & $1 \times 10^{-4}$ & 500 \\
\bottomrule
\end{tabular}
\end{table}

\begin{figure}[h]
    \centering
    \includegraphics[width=1\linewidth]{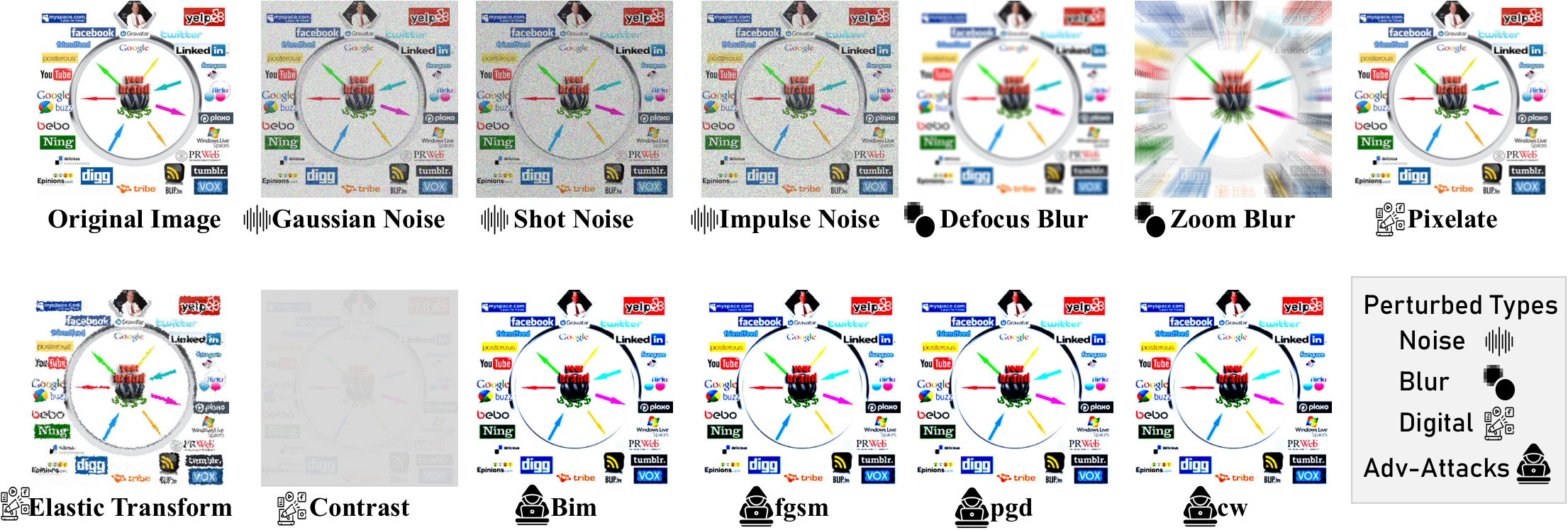}
    \caption{Example images under 8 natural image perturbations and 4 white-box adversarial attacks. The original image is taken from the Flickr30k dataset and shown on the top left.}
    \label{fig:image_perturbation}
\end{figure}

\subsection{Intersection over Union}
\label{iou}

In our experiments, we evaluate the accuracy of intermediate localization using the Intersection over Union (IoU) metric. Given two axis-aligned boxes $B_1=[x_1^{(1)},y_1^{(1)},x_2^{(1)},y_2^{(1)}]$ (prediction) 
and $B_2=[x_1^{(2)},y_1^{(2)},x_2^{(2)},y_2^{(2)}]$ (ground truth), IoU is defined as
\[
\mathrm{IoU}(B_1,B_2)=\frac{|B_1\cap B_2|}{|B_1\cup B_2|}.
\]

where $|B_1 \cap B_2|$ denotes the area of overlap between the two boxes, and $|B_1 \cup B_2|$ represents their union area. The overlap region is determined by taking the maximum of the top-left coordinates and the minimum of the bottom-right coordinates. The area of each box is then computed as the product of its width and height, and the union is given by the sum of both areas minus the intersection.

This implementation ensures that IoU ranges between 0 and 1, where 0 indicates no overlap and 1 denotes perfect alignment. In practice, a higher IoU signifies more accurate localization of predicted regions with respect to ground-truth annotations, while lower values reflect misalignment or degraded localization quality.

\subsection{Intermediate Perturbation Reuslts}
We conduct a supplementary experiment that perturbs not only the input image but also the intermediate local patches. This leads to a more pronounced performance drop, underscoring the sensitivity of \viscot to noise in intermediate components.

\label{intermediate-perturb}
\begin{table*}[h]
\centering
\renewcommand{\arraystretch}{1.2}
\caption{Answer accuracy (\%) of Visual CoT models under different perturbation positions: either applied on the global image only (“Global Only”) or both on the intermediate local crops (“Global and Local”). Experiments are conducted under severity level 5 across four datasets.}
\label{tab:midstep_injection}
\resizebox{\textwidth}{!}{
\begin{tabular}{c|c|c|cccccccccccc}
\toprule
\textbf{Dataset} & \textbf{Model} & \textbf{Perturb Location} &
\textbf{Gaussian} & \textbf{Shot} & \textbf{Impulse} & \textbf{Defocus} & \textbf{Zoom} & \textbf{Pixelate} & \textbf{Elastic} & \textbf{Contrast} &
\textbf{BIM} & \textbf{FGSM} & \textbf{PGD} & \textbf{C\&W} \\
\midrule

\multirow{4}{*}{\textbf{CUB}} 
& \multirow{2}{*}{\llavamodel } 
    & Global Only & 66.0 & 72.0 & 68.0 & 70.0 & 72.0 & 78.0 & 66.0 & 58.0 & 68.0 & 68.0 & 74.0 & 66.0 \\
&   & Global and Local  & \textbf{60.5} & \textbf{65.3} & \textbf{61.7} & \textbf{64.0} & \textbf{66.2} & \textbf{70.8} & \textbf{59.0} & \textbf{50.2} & \textbf{62.2} & \textbf{67.0} & \textbf{62.1} & \textbf{58.2} \\
\cmidrule(lr){2-15}
& \multirow{2}{*}{\viscotmodel } 
    & Global Only & 74.0 & 74.0 & 68.0 & 70.0 & 72.0 & 78.0 & 68.0 & 58.0 & 50.0 & 54.0 & 46.0 & 40.0 \\
&   & Global and Local  & \textbf{66.3} & \textbf{67.8} & \textbf{60.5} & \textbf{62.6} & \textbf{67.1} & \textbf{69.4} & \textbf{60.3} & \textbf{48.7} & \textbf{42.1} & \textbf{47.0} & \textbf{39.0} & \textbf{35.0} \\
\midrule

\multirow{4}{*}{\textbf{SROIE}} 
& \multirow{2}{*}{\llavamodel } 
    & Global Only & 24.0 & 27.0 & 29.0 & 29.9 & 30.0 & 30.0 & 30.0 & 21.5 & 10.4	&8.0	&\textbf{9.0}	&\textbf{8.6} \\
&   & Global and Local  & \textbf{20.3} & \textbf{23.5} & \textbf{25.0} & \textbf{26.2} & \textbf{25.8} & \textbf{27.2} & \textbf{25.1} & \textbf{17.0} & \textbf{14.6} & \textbf{16.2} & 12.5 & 9.3 \\
\cmidrule(lr){2-15}
& \multirow{2}{*}{\viscotmodel } 
    & Global Only & 26.0 & 29.0 & 35.0 & 28.0 & 25.0 & 31.0 & 30.0 & 12.0 & \textbf{6.0} &\textbf{12.0} &24.0 &\textbf{2.0} \\
&   & Global and Local  & \textbf{20.8} & \textbf{25.3} & \textbf{29.2} & \textbf{24.1} & \textbf{22.5} & \textbf{26.6} & \textbf{23.0} & \textbf{9.5} & 16.5 & 18.3 & \textbf{14.4} & 10.7 \\
\midrule

\multirow{4}{*}{\textbf{DocVQA}} 
& \multirow{2}{*}{\llavamodel } 
    & Global Only & 26.4 & 20.0 & 22.6 & 16.0 & 13.0 & 19.0 & 10.5 & 20.0 & 12.2	&16.0	&14.0	&12.4 \\
&   & Global and Local  & \textbf{21.2} & \textbf{16.7} & \textbf{18.0} & \textbf{13.4} & \textbf{10.6} & \textbf{15.8} & \textbf{8.1} & \textbf{15.5} & \textbf{11.6} & \textbf{13.0} & \textbf{10.0} & \textbf{7.2} \\
\cmidrule(lr){2-15}
& \multirow{2}{*}{\viscotmodel } 
    & Global Only & 29.2 & 24.1 & 26.0 & 15.3 & 11.5 & 29.0 & 18.7 & 19.6 & \textbf{12.1}	&\textbf{12.5}	&14.0	&\textbf{7.1}
 \\
&   & Global and Local  & \textbf{23.7} & \textbf{20.2} & \textbf{21.8} & \textbf{12.6} & \textbf{9.3} & \textbf{24.5} & \textbf{14.2} & \textbf{16.0} & 13.2 & 15.6 & \textbf{12.2} & 8.8 \\
\midrule

\multirow{4}{*}{\textbf{TextCaps}} 
& \multirow{2}{*}{\llavamodel } 
    & Global Only & 50.5 & 52.0 & 47.6 & 31.4 & 22.0 & 52.5 & 41.5 & 46.0 & 41.5	&\textbf{37.0}	&\textbf{19.0}	&44.0 \\
&   & Global and Local  & \textbf{44.3} & \textbf{47.1} & \textbf{42.0} & \textbf{27.0} & \textbf{18.2} & \textbf{46.8} & \textbf{35.0} & \textbf{38.4} & \textbf{34.1} & 37.8 & 30.1 & \textbf{25.3} \\
\cmidrule(lr){2-15}
& \multirow{2}{*}{\viscotmodel } 
    & Global Only & 51.0 & 59.0 & 54.0 & 61.0 & 58.0 & 63.0 & 56.0 & 54.0 & 41.0	&\textbf{37.5}	&\textbf{20.0}	&38.0 \\
&   & Global and Local  & \textbf{44.6} & \textbf{52.8} & \textbf{47.2} & \textbf{54.1} & \textbf{49.5} & \textbf{56.6} & \textbf{49.2} & \textbf{46.8} & \textbf{39.3} & 42.5 & 35.3 & \textbf{30.6} \\
\bottomrule
\end{tabular}
}
\end{table*}

\end{document}